\documentclass[journal]{IEEEtran}
\usepackage[font=small,labelfont=bf,justification=justified]{caption}
\usepackage{cite}
\ifCLASSOPTIONcompsoc
  \usepackage[caption=false,font=normalsize,labelfont=sf,textfont=sf]{subfig}
\else
 \usepackage[caption=false,font=footnotesize]{subfig}
\fi

\ifCLASSOPTIONcaptionsoff
  \usepackage[nomarkers]{endfloat}
 \let\MYoriglatexcaption\caption
 \renewcommand{\caption}[2][\relax]{\MYoriglatexcaption[#2]{#2}}
\fi

\usepackage{color}
\usepackage[table]{xcolor}
\definecolor{lightgray}{gray}{0.9}
\usepackage{multirow}
\usepackage{ctable}
\definecolor{myred}{RGB}{224, 64, 64}
\definecolor{mygreen}{RGB}{0, 176, 80}

\begin{document}
%
\title{ME R-CNN: Multi-Expert R-CNN for Object Detection}

%

\author{Hyungtae~Lee,~\IEEEmembership{Member,~IEEE,}
        Sungmin~Eum,~\IEEEmembership{Member,~IEEE,}
        and~Heesung~Kwon,~\IEEEmembership{Senior Member,~IEEE}
\thanks{Manuscript received July 9, 2018; revised April 15, 2019 and June 14, 2019; accepted August 27, 2019.}
\thanks{H. Lee and H. Kwon are with the Intelligent Perception Branch, the Computational \& Information Sciences Directorate (CISD), Army Research Laboratory, Adelphi, MD, 20783 USA (e-mail: \{hyungtae.lee,~heesung.kwon\}.civ@army.mil).}
\thanks{S. Eum is with Booz Allen Hamilton Inc., McLean, VA, 22102 USA and with the Intelligent Perception Branch, the Computational \& Information Sciences Directorate (CISD), Army Research Laboratory, Adelphi, MD, 20783 USA (e-mail: eum\_sungmin@bah.com).}
\thanks{© 2022 IEEE. Personal use of this material is permitted. Permission from IEEE must be obtained for all other uses, in any current or future media, including reprinting/republishing this material for advertising or promotional purposes, creating new collective works, for resale or redistribution to servers or lists, or reuse of any copyrighted component of this work in other works.}}

%
%

\markboth{IEEE Transactions on Image Processing}
{Shell \MakeLowercase{\textit{et al.}}: Bare Demo of IEEEtran.cls for IEEE Journals}
%



\maketitle

\begin{abstract}

We introduce Multi-Expert Region-based Convolutional Neural Network (ME R-CNN) which is equipped with multiple experts (ME) where each expert is learned to process a certain type of regions of interest (RoIs). This architecture better captures the appearance variations of the RoIs caused by different shapes, poses, and viewing angles. In order to direct each RoI to the appropriate expert, we devise a novel ``learnable'' network, which we call, expert assignment network (EAN). EAN automatically learns the optimal RoI-expert relationship even without any supervision of expert assignment. As the major components of ME R-CNN, ME and EAN, are mutually affecting each other while tied to a shared network, neither an alternating nor a naive end-to-end optimization is likely to fail. To address this problem, we introduce a practical training strategy which is tailored to optimize ME, EAN, and the shared network in an end-to-end fashion. We show that both of the architectures provide considerable performance increase over the baselines on PASCAL VOC 07, 12, and MS COCO datasets.

\end{abstract}


%
\IEEEpeerreviewmaketitle

\section{Introduction}
\label{sec:intro}

\IEEEPARstart{I}{n}  general, object detection uses distinctive shape patterns as evidence to find the object-of-interest in an image.  Object detection models are trained on these shape patterns that are commonly shown within the same object categories yet discriminative among the different categories. However, it is quite burdensome for a single model to accurately identify all the appearances since object appearances greatly vary according to fundamental object shape priors (e.g., airplane vs. person) as well as different object poses and viewing angles (e.g., a person lying down vs. standing upright).  Therefore, conventional object detection methods often use mixture of experts, each expert associated only with the corresponding shape patterns, in order to better capture large variations of object appearance ~\cite{FKhanCVPR2012,TMalisiewiczICCV2011,PFelzenswalbPAMI2010}.

\begin{figure}[t]
\begin{minipage}[b]{\linewidth}
  \begin{center}
  \centerline{\includegraphics[width=0.9\textwidth,trim=25mm 14mm 30mm 6mm]{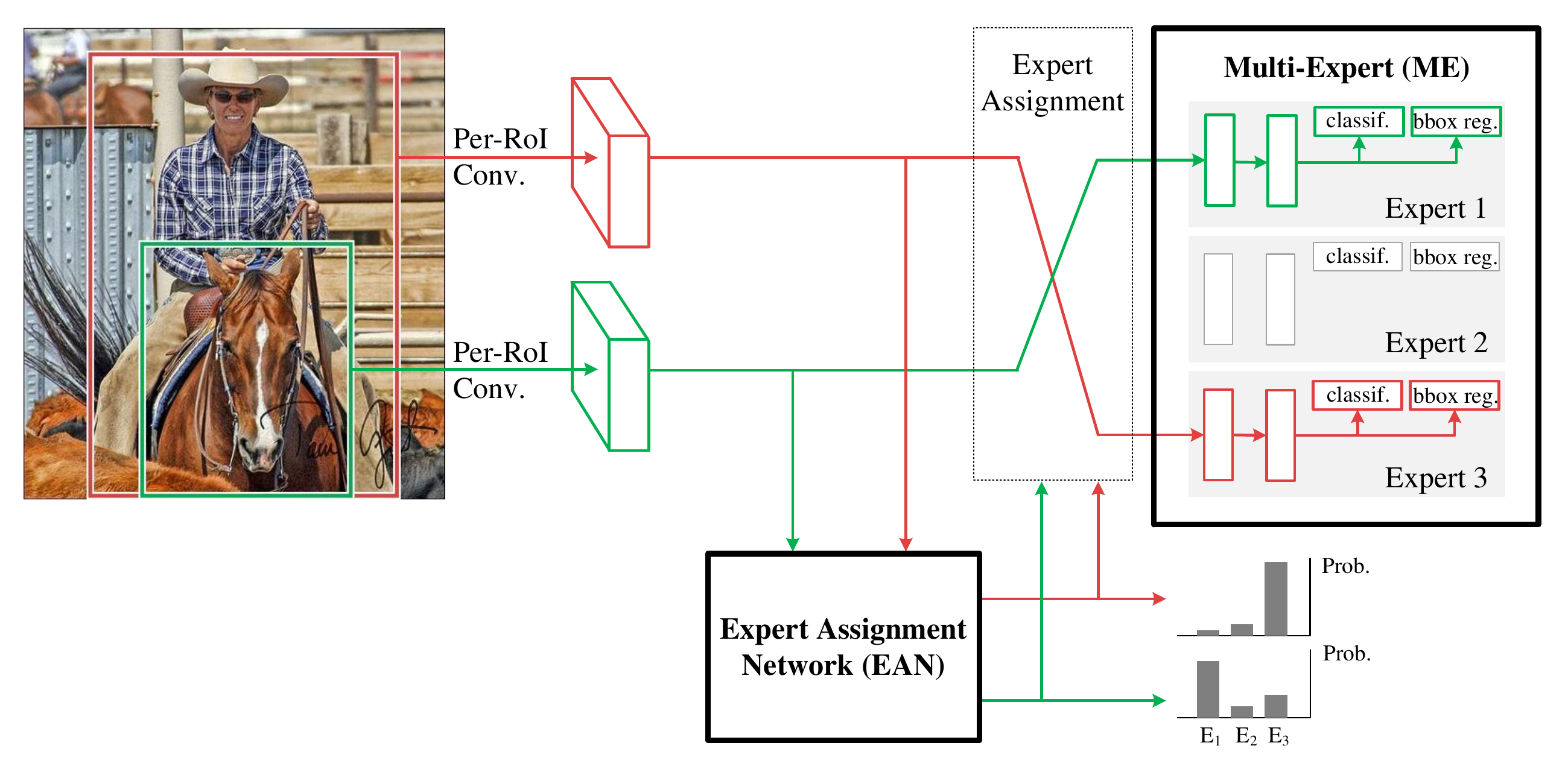}}
  \end{center}
\end{minipage}
\caption{{\bf ME R-CNN.} ME R-CNN adopts ``multi-expert (ME)'' to allow different streamlines for processing different RoIs. The ``expert assignment network (EAN)'' is built into the architecture to select the optimal streamline for each RoI. EAN is designed to output the probabilities for the RoI-expert relationships which guides the expert assignment.}
\label{fig:MERCNN}
\end{figure}

In this paper, we introduce a novel convolutional neural network (CNN)-based approach for object detection, referred to as ME R-CNN, which adopts multiple experts (ME). The ME R-CNN inherits the architecture of the region-based CNN (R-CNN)~\cite{RGirshickCVPR2014,RGirshickICCV2015,SRenNIPS2015,RGirshickPAMI2016,JDaiNIPS2016,SRenPAMI2017,KHeICCV2017,KHePAMI2018} which uses a single stream pipeline for processing each region-of-interest (RoI). However, unlike these approaches, the ME R-CNN is equipped with multiple stream pipelines, where one of the pipelines becomes an ``expert'' for processing a certain type of RoIs. To designate incoming RoIs to the appropriate experts, we construct a novel network called Expert Assignment Network (EAN). Figure \ref{fig:MERCNN} depicts the conceptual mechanism of the ME R-CNN which contains ME and EAN components.

The EAN is a ``learnable'' network which is trained to capture the RoI-expert relationship. It is designed to output a vector which indicates the matching probability for selecting one of the experts for a given RoI. The EAN consists of a convolutional, average pooling, and a fully connected layer. In the training scheme, the EAN is learned to choose the expert with minimum expert loss. The expert loss is defined as the sum of the losses for object classification and bounding box regression. These two losses were introduced in \cite{RGirshickICCV2015,SRenNIPS2015} to optimize the R-CNNs.

Training ME R-CNN is very challenging because: (1) ME and EAN components mutually affect each other (i.e., EAN training labels are defined based on the expert loss in ME, while EAN distributes the training samples to ME), (2) they are both derived from a shared network. To deal with (1), it is natural to use an alternating optimization strategy to co-train two mutually affecting tasks. However, this approach is likely to fail since the weights in the shared network optimized with respect to ME would no longer be in sync with EAN, and vice versa because of (2). To cope with these structural issues, it is necessary to come up with an end-to-end approach where all three components (ME, EAN, and the shared network) are optimized together.

However, there is a barrier which hinders the joint training of all three components. During the joint training, the RoI-expert relationships which are dependent upon the ME losses will be altered continuously, thus providing inconsistent (i.e., severely fluctuating) expert labels for the training of EAN. Therefore, we have devised an approach to go around this issue by adding a step where EAN and ME are learned separately to gain stability before proceeding into the joint learning stage. In this step, ME weights are firstly learned along with the shared network and then the RoI-expert relationships are learned by the EAN based on the pre-trained ME. Adding this ME/EAN initialization step was found to be useful in providing relatively consistent expert labels for EAN training, thus effectively assisting the joint optimization step which follows the initialization step. Note that, as expert labels for the RoIs do not exist at the time of initializing the ME, we have provided hard-coded labels for the RoIs which will be accounted for in the following paragraph. 

On top of advancing the object detection performance via employing multiple experts, ME R-CNN optimization can also be viewed as investigating the RoI-expert relationships in an unsupervised fashion (i.e., clustering) because no expert labels are ever provided as ground truth. Generally, clustering results highly depend on how initial cluster labels are assigned ~\cite{YSunPRL2002}. We have observed that the ME R-CNN learning is also highly sensitive to the presetting of the initial expert labels. When the expert labels are assigned by randomly initialized ME weights, most RoIs resulted in having assigned to one single expert after few training iterations, thus causing a failure in training. To avoid such extremely biased assignment, we also have tried forcing the RoIs to evenly be distributed to the experts for every training iteration, which also was found to be ineffective. Meanwhile, promising results were shown when initial assignments for the RoIs were done according to their aspect ratios.

We use Fast R-CNN \cite{RGirshickICCV2015} and Faster R-CNN \cite{SRenNIPS2015} for drawing up the baseline architecture of ME R-CNN. The Fast/Faster R-CNN architecture is composed of a network which intakes and processes holistic images (per-image network) which is followed by another network responsible for processing the RoIs (per-RoI network). For Faster R-CNN, the RoIs are generated by the region proposal network (RPN) while Fast R-CNN employs additional selective search process to obtain a set of RoIs. While ME R-CNN directly inherits the per-image network portion (and the RPN from the Faster R-CNN), the latter portion (per-RoI network) is replaced by our novel components: ME and EAN. We verified that ME R-CNN can consistently provide considerable performance boost over the baseline approaches in PASCAL VOC 07, 12, and MS COCO datasets.

The contributions of the proposed ME R-CNN can be summarized as follows:
\begin{enumerate}
    \item Introduction of ME R-CNN adopting ``multiple experts (ME)'' to better capture variations of the object appearance.
    
    \item Introduction of the EAN which can ``learn'' the RoI-expert relationship.
    
    \item Introduction of a practical training strategy to co-learn ME and EAN with a shared network in an end-to-end fashion.
    
    \item Considerable performance boost over the baselines on benchmark datasets.
\end{enumerate}

In Section~\ref{sec:rel_works}, we list out previous relevant literature and briefly introduce the innovative aspects of our approach. ME R-CNN architecture and its optimization strategy are described in Section~\ref{sec:MERCNN} and \ref{sec:learning}, respectively. Evaluation results are provided in Section~\ref{sec:experimentalEval}. We suggest future works in Section~\ref{sec:future} and provide the conclusion in Section~\ref{sec:Conclusion}.

\section{Related Works}
\label{sec:rel_works}

\subsection{Object Detection.}
Object detection is one of the most challenging tasks in computer vision.  Prior to the introduction of CNNs, non-CNN based object detection approaches, such as HOG-SVM (Histogram of Oriented Gradient - Support Vector Machines), DPM (Deformable Part Models), etc., were widely used for classifying RoIs into corresponding object categories~\cite{NDalalCVPR05,PFelzenswalbPAMI2010,FKhanCVPR2012,TMalisiewiczICCV2011}.  Within the past several years, multiple attempts have been made to use CNNs for object detection. Prominent methods among them are R-CNN~\cite{RGirshickCVPR2014} and its descendants~\cite{JDaiNIPS2016,RGirshickICCV2015,KHeICCV2017,KHeECCV2014,SRenNIPS2015} that provided the state-of-the-art performance for both localization accuracy and speed.

Although having achieved the top-notch performance, R-CNNs have not yet exploited some of the effective strategies which conventional object detection methods commonly use for boosting the performance. While the R-CNNs rely on heuristics to select hard negative examples, Shrivastava et al.~\cite{AShrivastavaCVPR2016} and Wang et al.~\cite{XWangCVPR2017} used the online hard example mining (OHEM) to automatically select hard examples with high optimization loss in every iteration of training. These approaches were motivated by the offline bootstrapping idea for training a classical object detection method \cite{NDalalCVPR05}.

Motivated by their successful practice, we focus on adding another conventional, yet effective, ``multi-expert'' flavor to the R-CNN architecture. Felzenswalb et al. \cite{PFelzenswalbPAMI2010} and Malisiewicz et al. \cite{TMalisiewiczICCV2011} have shown that employing multiple classifiers for the object detection task brings increase in performance.

\subsection{Mixture-of-Experts Models.} Multiple experts embedded in the proposed ME R-CNN is based on the concept of mixture-of-experts models.  The mixture-of-experts model is used to better estimate the probability distribution of a composite data with large variation (e.g., Gaussian mixture model~\cite{KYiCVPRW2013}).  In the image domain, object appearances can also show large variations according to their shapes, poses, and viewing angles.  Felzenswalb et al.~\cite{PFelzenswalbPAMI2010} nicely illustrates the importance of using a mixture of models by presenting two models, each of which captures the appearance of the front and the side view of a bicycle.  Accordingly, many approaches ~\cite{PFelzenswalbPAMI2010,EBernsteinCVPR2015,HSchneidermanCVPR2000} have shown that using the mixture-of-experts model for advanced object detection is very effective.  

\begin{figure}[t]
\begin{minipage}[b]{1.0\linewidth}
  \centering
  \centerline{\includegraphics[width=\textwidth,trim=5mm 5mm 5mm 5mm,clip]{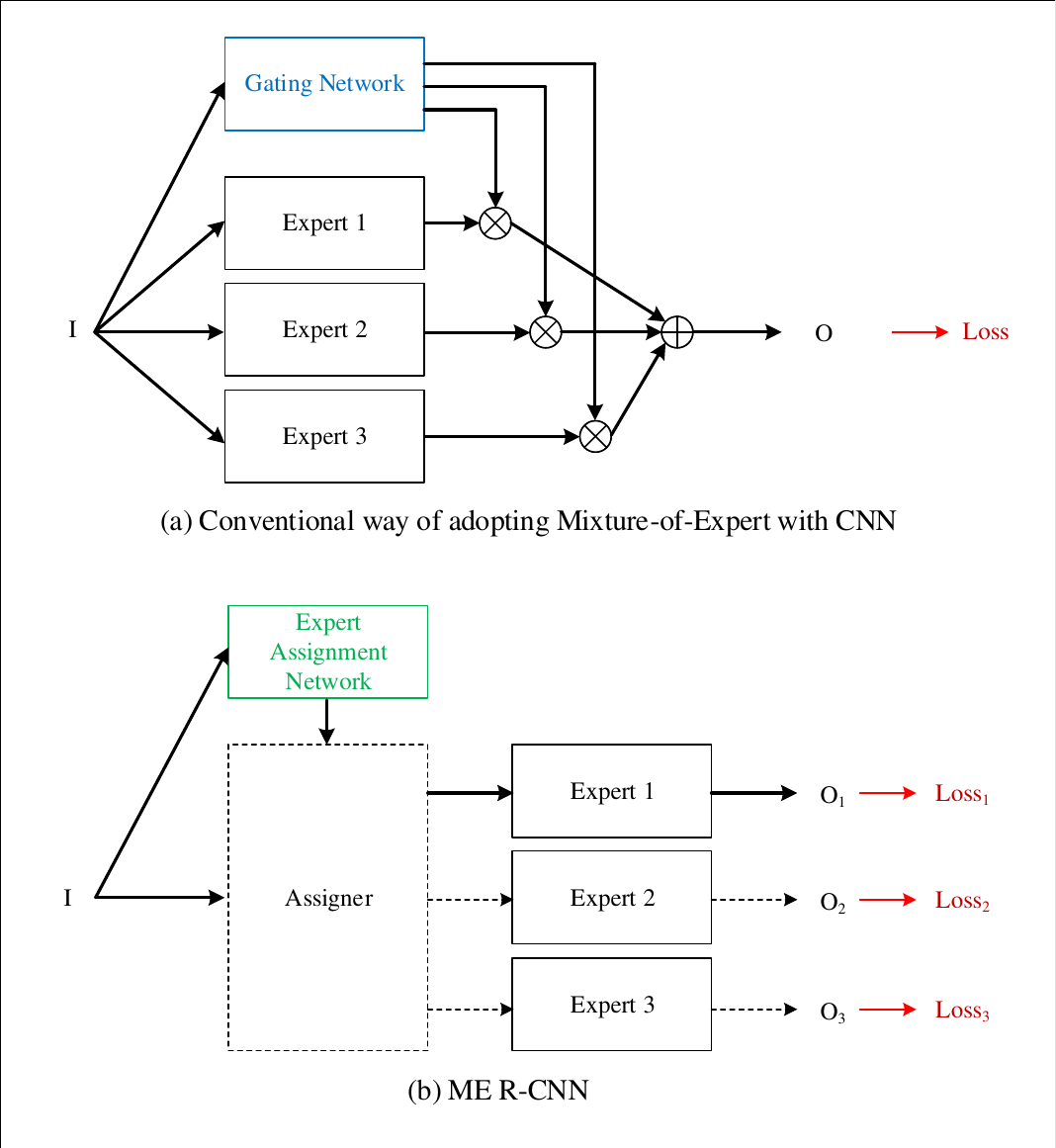}}
\end{minipage}
\caption{{\bf Conventional way of adopting mixture-of-expert with CNN vs. ME R-CNN.} Thick lines indicate the computational flow where one input has to go through. Only one of the three experts are activated in (b) whereas all the experts constantly have to be active to process one single example. I and O denote input and output, respectively. Assigner (dashed box) in ME R-CNN is not involved in training.}
\label{fig:mixture-of-expert}
\end{figure}

Recently, there have been several attempts to adopt the mixture-of-experts model in CNN-based recognition approach~\cite{AVermaICCVW2015,RRastiPR2017,SKumagaiArXiv2017,SGrossCVPR2017,RAljundiCVPR2017,AGarcisMartSensors2018}. Figure~\ref{fig:mixture-of-expert} shows how our way (ME R-CNN) of adopting mixture-of-experts in a CNN architecture is different from the conventional approach. In the conventional approaches (Figure \ref{fig:mixture-of-expert} (a)), all the experts are involved in processing each input while the gating network provides different weights to adjust the outputs of the experts. On the other hand, with ME R-CNN (Figure \ref{fig:mixture-of-expert} (b)), only one expert assigned by the EAN is activated, which makes it more efficient in terms of computation. Conventionally, all the experts are optimized with a same objective loss, but each expert in the ME R-CNN is optimized separately with its own loss in order to have unique expertise. To ensure that the overall performance does not suffer because of using only one expert at a time (instead of utilizing multiple experts each time), all the experts in the ME R-CNN architecture are trained concurrently in order to promote having experts with complimentary roles.

\subsection{Going Wider with CNN.}
One of the major innovations introduced in ME R-CNN is that the network has expanded in width, where the network width refers to the number of nodes in each layer. This is to equip the network with multiple number of specialized experts to better capture variations of object appearance. There have already been several attempts where the width of CNN architecture was expanded. Krizhevsky et al.~\cite{AKrizhevskyNIPS2012} splits each layer into two parallel layers in order to fully use two GPUs in a parallel fashion. 
Girshick~\cite{RGirshickICCV2015} appended two parallel layers with different functionalities at the end of the network, where the two layers are working for object category classification and bounding box regression, respectively. Szegedy et al.~\cite{CSzegedyCVPR2015} uses the inception module which employs multiple parallel layers in order to make use of dense sets of different sized convolutional filters. Xie et al.~\cite{SXieCVPR2017} modified the residual network structure~\cite{KHeCVPR2016} by replacing each residual module with multiple parallel sets of layers (i.e., branches). They referred to the number of branches as ``cardinality'' and show that increased cardinality of the network enhances the image classification performance. Several other approaches \cite{JDaiCVPR2016,SEumICIP2017,IKokkinosCVPR2017,HLeeICASSP2018,GGkioxariCVPR2018,HLeeIGARSS2018,HLeeICASSP2019,HLeeArXiv2019,HLeeIGARSS2019} also introduced widened networks for the task of co-learning multiple tasks in a single framework. 
Wang et al.~\cite{YWangCVPR2017} introduces the way of increasing model capacity such as depth or width during finetuning. ME R-CNN also enhances object detection accuracy by increasing the model width during finetuning.

\subsection{Unsupervised Learning (Clustering)}
Training ME R-CNN can be viewed as a type of a clustering approach because no expert labels are available at the time of training. For the cases where labels are not provided for a classification problem, many clustering methods are used such as K-means clustering~\cite{HRalambondrainyPRL1995,YSunPRL2002,AJainPRL2010}, mean-shift clustering~\cite{YChengPAMI1995,DComaniciuPAMI2002,LWangIP2015}, density-based spatial clustering~\cite{MEsterKDD1996,JHouIP2016,JShenIP2016}, expectation-maximization (EM) clustering~\cite{ADempsterJRSS1977,SSanjayGopalIP1998,HHongIP2009}, agglomerative hierarchical clustering~\cite{PFrantiPAMI2006,KHanASLP2008}. Most clustering methods start off the process with a label initialization step (e.g., random assignment) where initial cluster labels are assigned to each of the samples. How the labels are assigned initially may bring a significant impact in terms of clustering performance. For learning the ME R-CNN, we initially assigned the RoIs to multiple experts according to their aspect ratios. This initialization approach was found to be effective in avoiding the optimization divergence.

\section{ME R-CNN}
\label{sec:MERCNN}

\begin{figure*}[t]
\begin{minipage}[b]{1.0\linewidth}
  \centering
  \centerline{\includegraphics[width=\textwidth,trim=5mm 3mm 7mm 7mm]{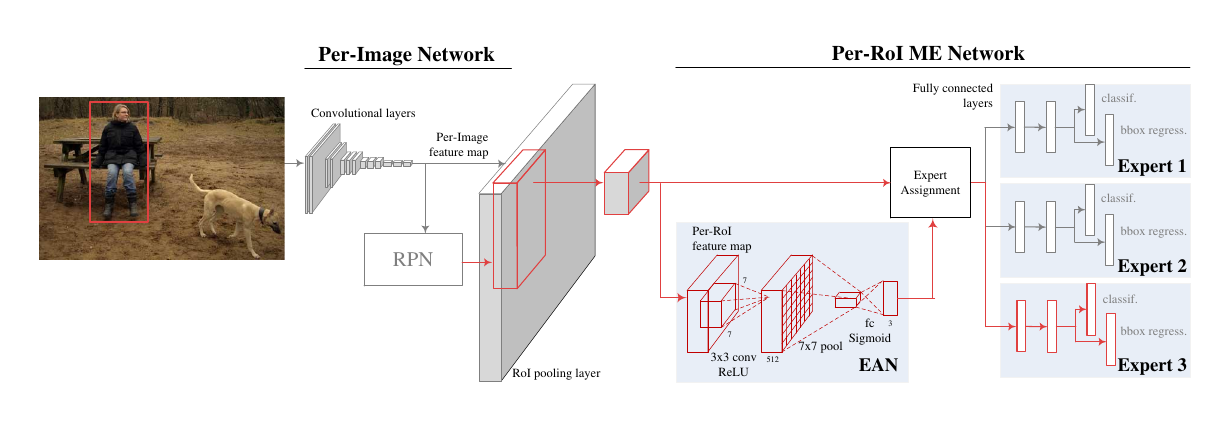}}
\end{minipage}
\caption{{\bf ME R-CNN architecture.} One example of a RoI-to-expert assignment is shown in \textcolor{myred}{red} arrows.}
\label{fig:me}
\end{figure*}

In this section, we first introduce Faster R-CNN \cite{SRenNIPS2015} as ME R-CNN inherits its structural backbone. Then the architectural components (ME and EAN) unique to ME R-CNN are elaborated in the following subsections. Lastly, we briefly describe how the network performs the task of object detection. The overall architecture of ME R-CNN is depicted in Figure \ref{fig:me}. 

\subsection{Faster R-CNN}
\label{ssec:fasterRCNN}

Faster R-CNN consists of a {\bf Per-Image Network} and a {\bf Per-RoI Network}. The {\bf Per-Image Network} can be divided into two parts: a set of convolutional layers (Conv-L) and a region proposal network (RPN). When an input image goes through the Conv-L, a per-image convolutional feature map is generated which is then fed into the RPN. The RPN is used to provide a set of hypothetical regions of interest (RoIs) for potential object regions. 

For all the RoIs from the RPN, RoI pooling layer crops out the corresponding regions from the per-image feature map. Each of these cropped-out feature maps (per-RoI feature maps) are max pooled to have a fixed size output. The output size is set to match the input size of the first fully-connected layer of the predefined CNN (e.g., 7$\times$7 for VGG16~\cite{KSimonyanICLR2015}).

All of these per-RoI feature maps are then fed into the {\bf Per-RoI Network}. At the end of this network, two sibling layers are present for object classification and bounding box regression. Object classification and bounding box regression are optimized using softmax classification loss and smooth $L_{1}$ loss, respectively.

\subsection{Per-RoI ME Network}
\label{ssec:MERCNN}

ME R-CNN resembles the overall architecture of Faster R-CNN as it contains both {\bf Per-Image Network} and {\bf Per-RoI Network}. {\bf Per-Image Network} of ME R-CNN inherits all the components from that of the Faster R-CNN. However, {\bf Per-RoI Network} portion is redesigned to fit the need of performing multi-expert supported object detection, and therefore, renamed as {\bf Per-RoI ME Network}. There are multiple stream pipelines built into the {\bf Per-RoI ME Network}, and each stream carries equivalent components in the {\bf Per-RoI Network} of Faster R-CNN. Each of the stream, known as an ``expert'', is responsible for processing a certain type of RoIs. To guide the RoIs to their best matching experts, we have constructed a network called Expert Assignment Network (EAN). Each expert is connected to its two loss functions for object classification and bounding box regression. Although our design does not constrain the number of experts, we have exploited three experts to be used for the following experiments and illustrations.

\subsection{Expert Assignment Network (EAN)}
\label{ssec:EAN}

For each RoI, its associated per-RoI feature map is fed into one of the three experts which is assigned by the EAN which is designed to ``learn'' the RoI-expert relationship. The EAN consists of two learnable layers, one convolutional and one fully connected layer. The input to the EAN are the per-RoI feature maps. When using VGG16 as the baseline architecture, the size of a feature map is 7$\times$7$\times$512. The convolutional layer employs 512 kernels (3$\times$3) with 1 stride and 1 padding. Then ReLU is applied which is followed by 7$\times$7 max-pooling generating a 1$\times$1$\times$512 output. A fully connected layer then takes this and generates 3 dimensional output, where each bin indicates the score for the corresponding expert. To allow the EAN to be able to select more than one expert for training, a binary sigmoid function is applied to the output. Architecture of the EAN is depicted in Figure \ref{fig:me}.

Assume that $f(x,W_{EAN})$ is the function which computes the output of the EAN with weight $W_{EAN}$ when given per-RoI feature map $x$. Note that we denote each entry of the EAN output as $f^{(e)}$ which indicates the assignment probability for the expert $e$. The expert assignment process which is being carried out by the EAN can then be formularized as: 
\begin{equation}
    e^{*} = \arg\max_{e\in\{E_1,E_2,E_3\}}{f^{(e)}(x, W_{EAN})}.
\end{equation}

EAN weights are optimized by minimizing the loss $L_{EAN}(\cdot)$ which intakes the EAN output generated by $f(\cdot)$ and the expert label vector \textbf{y} as shown below:

\begin{equation}
    W_{EAN}^{*} = \arg\min_{W_{EAN}}{L_{EAN}{(f(x,W_{EAN}),\textbf{y})}}
\label{eq:EANoptimization}
\end{equation}

The expert label vector \textbf{y} is constructed by concatenating the expert labels as $\textbf{y} = [y_{E_1}, y_{E_2}, y_{E_3}]$. Since the purpose of EAN is to find an expert which best performs in terms of object detection, each expert label $y_{e}$ is defined based on the expert loss $L_{e}$ as shown below:

\begin{equation}
\setlength{\tabcolsep}{2pt}
\begin{tabular}{lll}
\multirow{2}{*}{$y_e$ = \Big\{} & 1 & ~if $L_e(\{x,y_{obj}\},W_e)\leq \tau$ \\
& 0 & ~otherwise,
\end{tabular}
\label{eq:ye}
\end{equation}
where $y_{obj}$, $W_{e}$, and $\tau$ denote, respectively, object category label, expert weight, and assignment threshold. $L_{e}$, which is the expert loss for expert $e$, is computed as the sum of corresponding softmax classification loss and smooth $L_{1}$ bounding box regression loss. The assignment threshold $\tau$ is defined by the mean value of all the expert losses for each per-RoI feature map $x$.

\subsection{Object Detection}
\label{ssec:test}

In testing, a per-image convolutional map is generated by feeding an input image through the convolutional layers. Using this map and the RoIs provided by the RPN, per-RoI feature maps are acquired which all are then fed into the EAN. Each per-RoI feature map is sent to one of the experts which corresponds to the maximum EAN score. Note that, EAN is trained with the binary labels for each RoI-expert assignment, which allows having each RoI being assigned to more than one expert at a time. This training strategy eventually prevents performance degradation even when a RoI is not assigned to the most desirable expert in the testing phase. This can be intuitively seen as preparing more than one experts which can properly function with confusing RoIs. In testing, each RoI is assigned to only one expert in order to achieve the same level of computational complexity as single expert model.

ME R-CNN outputs three sets of detection results per image, i.e., bounding boxes and their scores, from three different experts.  The bounding boxes are refined by incorporating the output of bounding box regression layers.  We combine these three sets of detection results and apply non-maximum suppression (NMS) with overlap criteria of 0.3 for each object category.

Note that, computational load which is required to process each RoI using ME R-CNN is comparable to the case when Faster R-CNN is used because only one expert is activated for an RoI in ME R-CNN while EAN adds negligible computational cost. Therefore, as long as the same number of RoIs are used, computational costs for the ME R-CNN and Faster R-CNN are highly similar.

\section{Learning ME R-CNN}
\label{sec:learning}

We base our training strategy on the pragmatic ``4-step alternating optimization" devised by Ren et al. \cite{SRenNIPS2015} but modify it to fit the need of the components in the ME R-CNN. In the first step, we train the RPN. As RPN takes the output of the Conv-L, we use the ImageNet-pre-trained model to initialize the Conv-L for this step. In the second step, we train ME, EAN, and Conv-L using sets of region proposals generated by the step-1 RPN. Conv-L is again initialized by the ImageNet-pre-trained model. In the third step, we re-train the RPN to be aligned with the newly trained Conv-L from step-2. Finally, we update ME and EAN with step-2 Conv-L and step-3 RPN. In this step, ME and EAN weights are initialized by step-2 ME and EAN, respectively. Note that in step-3 and step-4, Conv-L is fixed.

Overall training strategy for the ME R-CNN is listed out in Algorithm 1. In the following subsections, we elaborate on step-2 and step-4 which are devised to train the components (ME, EAN, and Conv-L) unique to the ME R-CNN. For training the RPN in step-1 and step-3, we have followed the procedures introduced in \cite{SRenNIPS2015}.

\begin{table}[ht]
\setlength{\tabcolsep}{14.0pt}
\renewcommand{\arraystretch}{1.4}
\begin{center}
{\normalsize
\begin{tabular}{ll}
\specialrule{.15em}{.05em}{.05em} 
\multicolumn{2}{l}{{\bf Algorithm 1:} 4-step alternating algorithm} \\\specialrule{.15em}{.05em}{.05em}
1 & Train RPN \& Conv-L \\
& (Conv-L from 1 is no longer used hereafter.)\\
2 & Train ME, EAN, \& Conv-L \\
2a & ~~~~~~Initialize ME \& Conv-L \\
2b & ~~~~~~Initialize EAN \\
2c & ~~~~~~Co-train ME, EAN, \& Conv-L \\
3 & Train RPN \\
4 & Update ME \& EAN \\
4a & ~~~~~~Update EAN \\
4b & ~~~~~~Update ME
\\\specialrule{.15em}{.05em}{.05em} 
\end{tabular}
\label{alg:optimization}
}
\end{center}
\end{table}

\subsection{Step-1: Train RPN \& Conv-L}
\label{ssec:rpn_train_1}

RPN evaluates all spatial locations from the per-image convolutional feature map. Every location in the map is mapped to multiple windows which are predefined in sizes (e.g., 8$\times$16, 16$\times$8, 8$\times$8, etc). Each mapping reference is referred to as an ``anchor''. For PASCAL VOC, three scales and three aspect ratios, represented with 9 anchors, are considered to obtain region proposals via RPN. These anchors are 16$\times$8, 8$\times$8, 8$\times$16, 32$\times$16, 16$\times$16, 16$\times$32, 64$\times$32, 32$\times$32, and 32$\times$64. For MS COCO, four more anchors (8$\times$4, 4$\times$4, and 4$\times$8) are added to consider objects with extremely small size.

For training RPN, positive/negative examples are chosen from all possible windows according to the IOU (intersection-over-union) between the window bounding boxes and the groundtruth bounding boxes of any objects. Windows with IOUs larger than 0.7, are treated as positive training examples. Windows with IOUs between 0.7 and 0.3, are used as negative examples. RPN is trained using mini-batches, where each mini-batch consists of 128 positive and 128 negative examples.

The trained RPN is used to provide the region proposals in training ME, EAN \& Conv-L (step-2). In order to provide more accurate region proposals, Conv-L is also trained in step-1. However, the Conv-L weights learned in this step are no longer used afterwards.

\subsection{Step-2: Train ME, EAN, \& Conv-L}
\label{ssec:co-training}

\begin{figure}[t]
\begin{minipage}[b]{1.0\linewidth}
  \centering
  \centerline{\includegraphics[width=\textwidth,trim=7mm 4mm 7mm 7mm]{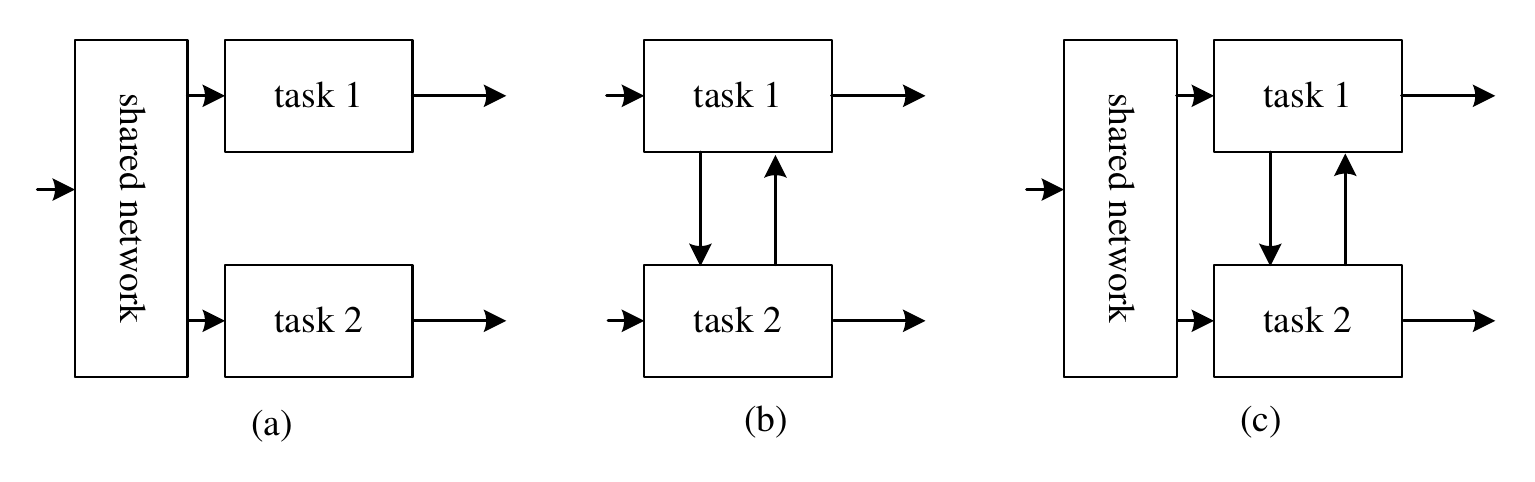}}
\end{minipage}
\caption{{\bf Network architecture comparison.} (a) Shared network-driven multiple task (b) Mutually-affecting multiple task (c) Shared network-driven, mutually-affecting multiple task (In ME R-CNN, shared network: Conv-L, tasks 1 \& 2: EAN \& ME) }
\label{fig:networkComparison}
\end{figure}

In this subsection, we introduce a novel way to simultaneously train the major components (ME, EAN, and Conv-L) in the ME R-CNN which are layed out differently when compared with other types of multi-task learning architectures. Many multi-task learning architectures \cite{JDaiCVPR2016,SEumICIP2017,IKokkinosCVPR2017} are designed so that the tasks are independently derived from the commonly shared network as shown in Figure \ref{fig:networkComparison}(a). Another stream of multi-task architectures do not require shared networks while having mutually-affecting tasks as depicted in Figure \ref{fig:networkComparison}(b). The Generative Adversarial Network (GAN) \cite{IGoodfellowNIPS2014} can be considered as a representative example of an architecture which contains mutually-affecting multiple tasks (`Generator' and `Discriminator'). Shared network-driven multiple task architectures can be trained by enforcing separate loss functions in an end-to-end fashion, while the architecture with mutually-affecting tasks are typically trained using an alternating optimization strategy as the module responsible for one task needs to be fixed to train the other.

Unlike the two types, the architecture of ME R-CNN can be viewed as having tasks (ME and EAN) derived from the shared network (Conv-L), and at the same time, mutually affecting each other. Figure \ref{fig:networkComparison}(c) illustrates the schematic architecture of ME, EAN, and Conv-L. This makes the training very challenging as neither of the previously mentioned training approaches can be applied directly. 

Attempting to train these modules using an alternating optimization strategy is far from reaching the optimal point and likely to fail. For instance, the shared network portion constantly changes every time task 1 is being optimized, but at the same time loses its sync with respect to task 2. The problem is that previously optimized task 2 is no longer optimized with respect to the shared network but still harmfully affects task 1.

To cope with these structural issues, all three components (Me, EAN, and Conv-L) need to be optimized together. However, when carrying out the joint learning, constantly changing ME provides inconsistent (severely fluctuating) expert labels, which adversely impacts EAN training. To gain stability before proceeding into the joint learning (Step-2c), we add an initialization step where ME weights along with Conv-L are learned (Step-2a) which is then followed by the EAN learning (Step-2b) in which the pre-trained ME weights are fixed. We have observed that these initialization steps were effective in providing relatively consistent expert labels for the EAN training and thus accommodating better grounds for the joint optimization as shown in Table~\ref{tab:co-learning}.\medskip

\begin{figure*}[t]
\begin{minipage}[b]{1.0\linewidth}
  \centering
  \centerline{\includegraphics[width=\textwidth,trim=7mm 4mm 7mm 7mm]{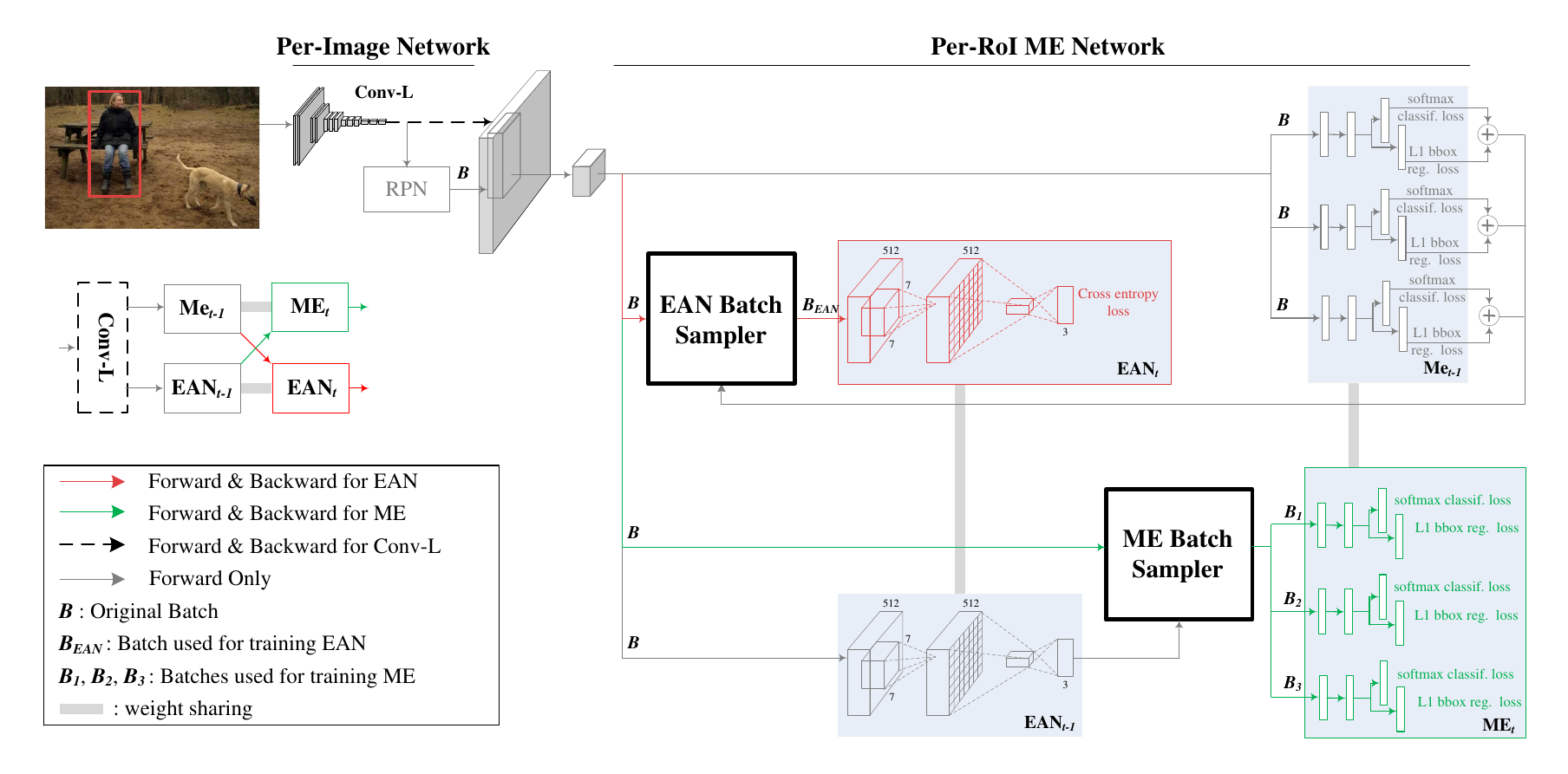}}
\end{minipage}
\caption{{\bf Protocol of co-training ME, EAN \& Conv-L.} A conceptual illustration of the entire flowchart is shown below the input image.}
\label{fig:co-trainingProtocol}
\end{figure*}

\noindent{\bf Initialize ME \& Conv-L.} When initializing ME and Conv-L independent of EAN (i.e., without any expert assignment information from EAN), temporarily exploiting RoI shape-based assignment criteria in place of EAN was found to be effective. Each RoI is labeled with a shape category chosen among horizontally elongated ({\bf H}), square-like ({\bf S}), or vertically elongated ({\bf V}) according to its aspect ratio. 

We denote $w$ and $h$ as the width and the height of an RoI. All RoIs satisfying $w>h$ are assigned to the {\bf H} category. All RoIs satisfying $w<2h$ and $w>\frac{1}{2}h$ are assigned to the {\bf S} category. Lastly, RoIs with $w<h$ are assigned to the {\bf V} category. Note that, under this RoI assignment criteria, an RoI can be categorized into more than one category. This is done to have multiple experts responsible for the RoIs which can be shared across the different categories while training the network. We trained three different experts using the RoIs assigned to {\bf H}, {\bf S}, and {\bf V} category, respectively.

To optimize the three experts, three batches are prepared for every iteration.  Each batch is built from two images, and each image contributes 64 randomly chosen RoIs.  For each expert, only the RoIs that match its associated shape category are selected for training.  Each RoI is labeled as a positive or negative example according to an IOU overlap criteria between the RoI and the groundtruth bounding box.  The RoIs having IOU overlap equal to or bigger than 0.5 are labeled as positive examples and the remaining ones are labeled as negative. For each batch, the ratio between the number of positive and negative examples is fixed as 1:3. Batch preparations for all of the training procedures in this paper (except $B$ in `Co-learn ME, EAN \& Conv-L') are equivalently done as described in `Initialize ME \& Conv-L'.

Conv-L and ME are each finetuned from the convolutional layers and the fully connected layers of the ImageNet-pretrained model, respectively. Three pairs of sibling layers (classification and bounding box regression) appended at the end of the expert streams are initialized as well. The classification layer weights are initialized by randomly selecting them according to Gaussian distribution with the mean of 0 and the standard deviation of 0.01.  For the bounding box regression layer, we initialized the weights randomly selected from Gaussian distribution with the mean and the standard deviation of 0 and 0.001, respectively. When finetuning Conv-L, we multiply 1/3 to the base learning rate because optimizing the layers in Conv-L is affected by all three streams of ME at each training iteration when back-propagation takes place. This scheme of multiplying 1/3 to the base learning rate in finetuning Conv-L is also used when we co-learn ME and EAN.\medskip

\noindent{\bf Initialize EAN.} The weights in EAN are learned according to Equation \ref{eq:EANoptimization} while fixing weights in Conv-L and ME. The fixed weights for Conv-L and ME are inherited from the results of `Initialize ME \& Conv-L'. Note that before learning EAN, the weights (both for convolutional and fully connected layers) are initialized by randomly selecting them according to Gaussian distribution with mean and standard deviation of 0 and 0.01, respectively. \\

\noindent{\bf Co-learn ME, EAN \& Conv-L.} The overall protocol of co-training ME, EAN \& Conv-L is depicted in Figure \ref{fig:co-trainingProtocol}. Let us assume that the weights for ME (same applies to EAN and Conv-L) learned in the previous and current iteration are denoted as ME$_{t-1}$ and ME$_{t}$, respectively.
Once RPN generates a set of region proposals, the whole set of corresponding per-RoI feature maps ($B$) is forward passed into ME$_{t-1}$ and EAN$_{t-1}$.

Forward passed output from ME$_{t-1}$ is then fed into the EAN Batch Sampler which outputs a downsized batch ($B_{EAN}$) which contains per-RoI feature maps selected based on the expert loss $L_{e}$. Note that $L_{e}$, which is the expert loss for expert $e$, is the sum of corresponding softmax classification loss and smooth $L_{1}$ bounding box regression loss. $B_{EAN}$ is then used to train EAN$_t$. The ground truth labels for the samples in the batch are defined according to Equation \ref{eq:ye}. 

In a similar manner, output of EAN$_{t-1}$ is fed into the ME Batch Sampler which generates equal-sized three sets of batches, $B_{1}$, $B_{2}$, and $B_{3}$ which will be used to train ME$_{t}$. In the ME Batch Sampler, each batch is generated by collecting the per-RoI feature maps and their associated object category labels which were tagged with 0.8 or higher EAN output probability. EAN output is a 3-dimensional vector where each entry indicates the matching probability for the corresponding expert among three.

Along with the training of EAN$_t$ and ME$_t$, Conv-L is concurrently trained via back-propagation. For the very first iteration of training, ME$_{0}$ and EAN$_{0}$ are used in place of ME$_{t-1}$ and EAN$_{t-1}$, respectively, which are acquired by previously mentioned `Initialize ME \& Conv-L' and `Initialize EAN'.

\subsection{Step-3: Train RPN}
\label{ssec:rpn_train_2}

In this step, RPN is finetuned from the model trained in step-1. Unlike step-1, conv-L weights are not updated in order to share them among RPN, EAN, and ME. All the details (i.e., hyper-parameters, anchors, training example labeling) are same to those used in step-1. Updated RPN generates the region proposals used in updating EAN and ME in step-4.

\subsection{Step-4: Update EAN \& ME}

We generate the region proposals using the RPN learned in step-3 to update the weights in EAN and ME. This is carried out to better align EAN and ME with respect to the newly trained RPN and maximize the performance. EAN weights are updated first while fixing ME weights, and then ME is updated by fixing EAN. This alternating update procedure for EAN and ME is feasible for this step since Conv-L is fixated with the previously learned weights.

\subsection{Joint Training of All Components}
4-step alternating optimization is a sub-optimal approach. As Ren et al.~\cite{SRenPAMI2017} recently provided a joint training strategy for learning the Faster R-CNN, we also have tried a single-step joint training of all the components (ME, EAN, and Conv-L) for our network. However, the jointly trained ME R-CNN did not perform well in terms of detection accuracy. As previously mentioned (Section \ref{sec:intro} and Section \ref{ssec:co-training}), jointly optimizing ME R-CNN in an end-to-end fashion is highly sensitive to the presetting of RoI-expert assignment. As our future work, we will seek to develop a more efficient and effective learning strategy for ME R-CNN.

\section{Experimental Evaluation}
\label{sec:experimentalEval}

In this section, we evaluate our method on VOC 2007, 2012 \cite{MEveringhamIJCV2015} as well as MS COCO \cite{TLinECCV2014} dataset.

\subsection{Experimental Setup}
\label{ssec:experimentalSetup}

\begin{table*}[t]
\setlength{\tabcolsep}{4.8pt}
\renewcommand{\arraystretch}{1.4}
\begin{center}
\begin{tabular}{c||c|c||c|c|c|c|c|c|c|c}
\specialrule{.15em}{.05em}{.05em}
\multirow{2}{*}{train set} & Fast R-CNN & ME R-CNN & \multicolumn{4}{c|}{Faster R-CNN} & \multicolumn{4}{c}{ME R-CNN} \\\cline{2-11}
& & 2a/2b/2c & 1 & 2 & 3 & 4 & 1 & 2a/2b/2c & 3 & 4a/4b \\\specialrule{.15em}{.05em}{.05em}
{\bf 07} & 40k/30k & 40k/30k & 80k/60k & 40k/30k & 80k/60k & 40k/30k & 80k/60k & 40k/30k & 80k/60k & 40k/30k \\
{\bf 12} & 40k/30k & 40k/30k  & 80k/60k & 40k/30k & 80k/60k & 40k/30k & 80k/60k & 40k/30k & 80k/60k & 40k/30k \\ 
{\bf 07+12} & 100k/75k & 100k/75k & 200k/150k & 100k/75k & 200k/150k & 100k/75k & 200k/150k & 100k/75k & 200k/150k & 100k/75k \\
{\bf 07++12} & 120k/90k & 120k/90k & 240k/180k & 120k/90k & 240k/180k & 120k/90k & 240k/180k & 120k/90k & 240k/180k & 120k/90k \\
{\bf coco train} & $\cdot$ & $\cdot$ & 320k/240k & 320k/240k & 320k/240k & 320k/240k & 320k/240k & 160k/120k & 320k/240k & 160k/120k \\
{\bf coco trainval} & $\cdot$ & $\cdot$ & 320k/240k & 320k/240k & 320k/240k & 320k/240k & 320k/240k & 160k/120k & 320k/240k & 160k/120k
\\\specialrule{.15em}{.05em}{.05em}

\end{tabular}
\end{center}
\vspace{-0.3cm}
\caption{{\bf Minibatch iterations and step sizes} of ME R-CNNs and their baseline architectures for different trainsets and training steps. Training set key: {\bf 07}: VOC07 trainval, {\bf 12}: VOC12 trainval, {\bf 07+12}: union of VOC07 trainval and VOC12 trainval, {\bf 07++12}: union of VOC07 trainval, VOC07 test, and VOC12 trainval, {\bf coco train}: MS COCO train, {\bf coco trainval}: MS COCO train and val.}
\label{tab:iteration}
\end{table*}

We use VGG16~\cite{KSimonyanICLR2015} or ResNet-101~\cite{KHeCVPR2016} as a predefined CNN for all experiments. We also use Fast R-CNN~\cite{RGirshickICCV2015} or Faster R-CNN~\cite{SRenNIPS2015} as our baseline architectures for the ME R-CNN. When Faster R-CNN is chosen as the base architecture for ME R-CNN, all 4-steps are carried out. However, when Fast R-CNN is used, single-step end-to-end optimization (i.e., step 2 of ME R-CNN optimization) is carried out as RPN-related training procedures are not required. For all the methods we have tested, we used stochastic gradient descent with a base learning rate of 0.001 and the weight decay of 0.1. As reported in Table \ref{tab:iteration}, the minibatch iterations and the step sizes were varied according to trainsets and training steps. The mini-batch iteration and step size of the ME R-CNN optimization is determined according to each of its baseline architecture.

For all evaluations, we use single-scale training/testing as in \cite{RGirshickICCV2015}, by setting the shorter side of the images to be 600 pixels. We have carried out all the experiments on Caffe framework~\cite{YJiaACMMM2014} with a Titan XP GPU.\medskip

\begin{figure}[t]
\begin{minipage}[b]{1.0\linewidth}
  \centering
  \centerline{\includegraphics[width=\textwidth,trim=12mm 7mm 12mm 7mm]{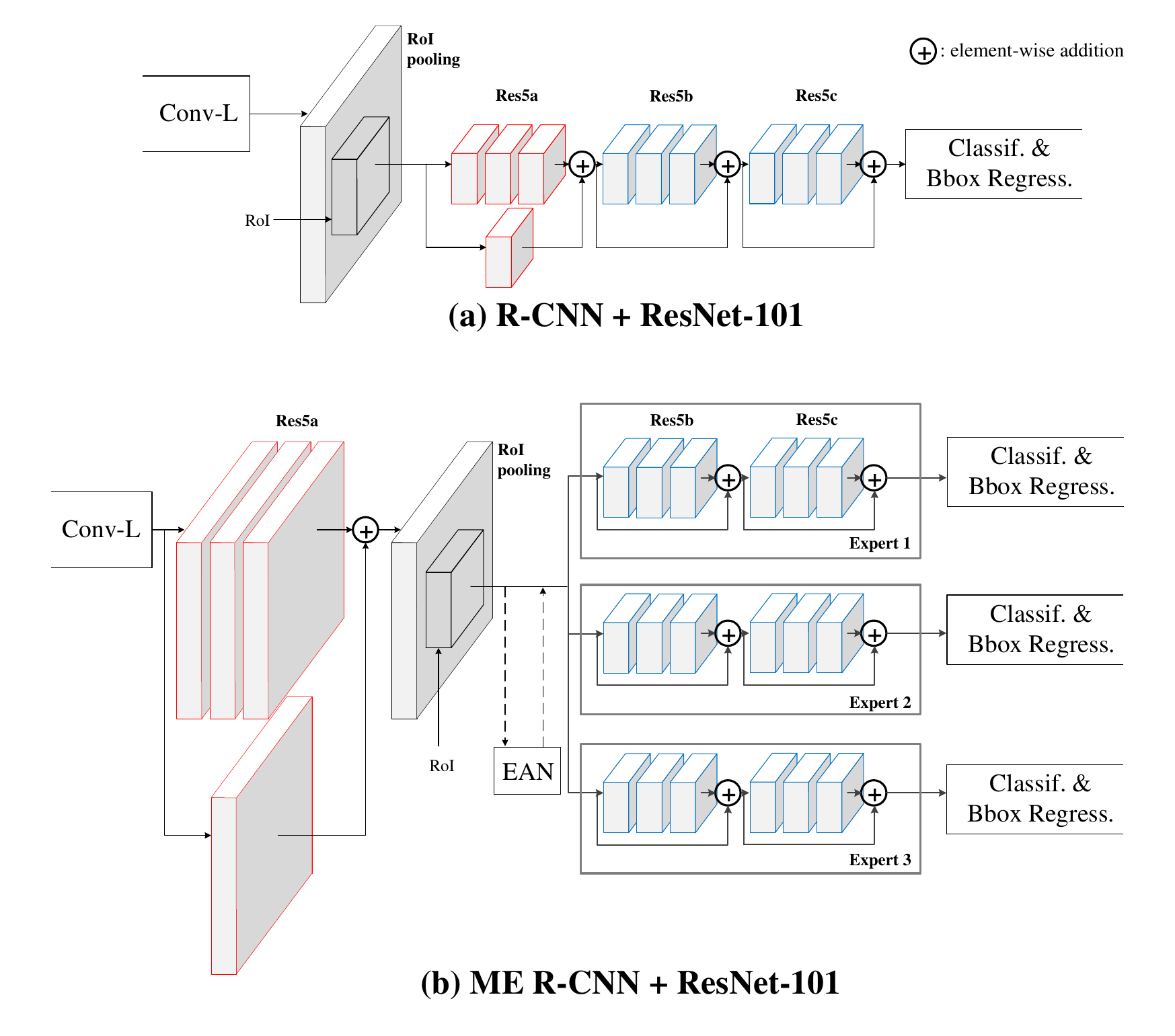}}
\end{minipage}
\caption{{\bf ME R-CNN adopting ResNet-101 architecture.} Per-RoI network of ResNet-101 consists of three residual modules (Res5a, Res5b, and Res5c). One can observe that the position of Res5a (red) with respect to the RoI pooling has changed.}
\label{fig:me2resnet}
\end{figure}

\noindent{\bf Adapting Multi-Expert into ResNet-101.} To use ResNet-101 as the backbone network for object detection, He et al.~\cite{KHeCVPR2016} exploited the last 10 convolutional layers of ResNet-101 to function as the per-RoI network as depicted in Figure \ref{fig:me2resnet}(a). Denote three residual modules consisting of the last 10 convolutional layers as Res5a, Res5b, and Res5c, respectively. In our case, as shown in Figure \ref{fig:me2resnet}(b), the last 6 convolutional layers (Res5b and Res5c) are used as the per-RoI multi-expert network and the first 4 convolutional layers (Res5a) are appended at the end of the per-image convolutional network due to the GPU memory limitation.

In ResNet-101, RoI pooling layer takes a per-image convolutional feature map and outputs 14$\times$14 per-RoI feature maps. Per-RoI feature map resolution is halved to 7$\times$7 during processing in Res5a. For our architecture, in order to maintain the resolution of the per-image feature map (i.e., an input of RoI pooling layer as well as the output of the Res5a), we do not use a resizing function of Res5a. Instead, we modify the output dimension of the RoI pooling layer as 7$\times$7 in order to preserve the following layers (Res5b and Res5c) which are inherited from ResNet-101.

We also reduce the batch size from 128 to 64 for training while taking more training iterations. This change is also considered in order to cope with the GPU memory limitation. 

We have made above modifications in order to adopt our ME component into an architecture which consists of Faster R-CNN and ResNet-101. Note that Faster R-CNN + ResNet-101 combination is widely used as the backbone architecture in state-of-the-art object detection approaches. (See Table \ref{tab:top_ranked_methods}.)

\subsection{VOC 2007 and 2012 Results}
\label{ssec:VOCResults}

\begin{table*}[t]
\setlength{\tabcolsep}{2.0pt}
\renewcommand{\arraystretch}{1.4}
\begin{center}
\begin{tabular}{l|c|c|ccccccccccccccccccccc}
\specialrule{.15em}{.05em}{.05em}
\multirow{2}{*}{method} & \multirow{2}{*}{trainset} & \multirow{2}{*}{mAP (\%)} & \multicolumn{20}{c}{category} \\
&&& aero & bike & bird & boat & bottle & bus & car & cat & chair & cow & table & dog & horse & mbike & persn & plant & sheep & sofa & train & tv \\\specialrule{.15em}{.05em}{.05em}
Fast R-CNN \cite{RGirshickICCV2015} & {\bf 07} & 66.9 & 74.5 & 78.3 & 69.2 & 53.2 & 36.6 & 77.3 & 78.2 & 82.0 & 40.7 & 72.7 & 67.9 & 79.6 & 79.2 & 73.0 & 69.0 & 30.1 & 65.4 & 70.2 & 75.8 & 65.8 \\
ME R-CNN & {\bf 07} & {\bf 69.0} & 70.6 & 78.9 & 68.2 & 55.8 & 44.3 & 80.9 & 78.2 & 84.6 & 44.4 & 76.5 & 70.4 & 80.6 & 81.5 & 76.6 & 70.8 & 35.1 & 66.4 & 69.9 & 76.8 & 69.5 \\\hline
Faster R-CNN \cite{SRenNIPS2015} & {\bf 07} & 69.9 & 70.0 & \underline{{\bf 80.6}} & 70.1 & 57.3 & 49.9 & 78.2 & 80.4 & 82.0 & 52.2 & 75.3 & 67.2 & 80.3 & 79.8 & 75.0 & 76.3 & 39.1 & 68.3 & 67.3 & 81.1 & 67.6 \\
ME R-CNN & {\bf 07} & {\bf 70.6} & 69.9 & 79.5 & 68.2 & 58.5 & 52.5 & 77.1 & 80.1 & 84.4 & 52.1 & 78.6 & 67.3 & 81.0 & 83.7 & 74.6 & 77.0 & 38.2 & 71.0 & 66.0 & 75.2 & \underline{{\bf 74.7}} \\\hline
Fast R-CNN \cite{RGirshickICCV2015} & {\bf 07+12} & 70.0 & 77.0 & 78.1 & 69.3 & 59.4 & 38.3 & 81.6 & 78.6 & 86.7 & 42.8 & 78.8 & 68.9 & 84.7 & 82.0 & 76.6 & 69.9 & 31.8 & 70.1 & 74.8 & 80.4 & 70.4 \\
ME R-CNN & {\bf 07+12} & {\bf 72.2} & \underline{{\bf 78.1}} & 78.9 & 69.4 & 61.3 & 44.5 & 84.8 & 81.7 & 87.6 & 50.7 & 80.1 & \underline{{\bf 70.6}} & 85.8 & \underline{{\bf 84.8}} & \underline{{\bf 78.7}} & 72.3 & 35.0 & 71.9 & \underline{{\bf 75.3}} & 79.9 & 72.7 \\\hline
Faster R-CNN \cite{SRenNIPS2015} & {\bf 07+12} & 73.2 & 76.5 & 79.0 & 70.9 & 65.5 & 52.1 & 83.1 & 84.7 & 86.4 & 52.0 & 81.9 & 65.7 & 84.8 & 84.6 & 77.5 & 76.7 & 38.8 & 73.6 & 73.9 & 83.0 & 72.6 \\
ME R-CNN & {\bf 07+12} & \underline{{\bf 75.8}} &  77.2 & 79.7 & \underline{{\bf 76.3}} & \underline{{\bf 67.0}} & \underline{{\bf 60.3}} & \underline{{\bf 86.0}} & \underline{{\bf 87.1}} & \underline{{\bf 88.6}} & \underline{{\bf 58.3}} & \underline{{\bf 83.8}} & 70.3 & \underline{{\bf 86.4}} & 84.7 & 78.4 & \underline{{\bf 78.4}} & \underline{{\bf 45.1}} & \underline{{\bf 76.0}} & 73.8 & \underline{{\bf 83.6}} & 74.6 \\\specialrule{.15em}{.05em}{.05em}
\end{tabular}
\end{center}
\vspace{-0.3cm}
\caption{{\bf VOC 2007 detection accuracy.} All methods use VGG16.}
\label{tab:voc07}
\end{table*}

\begin{table*}[t]
\setlength{\tabcolsep}{2.0pt}
\renewcommand{\arraystretch}{1.4}
\begin{center}
\begin{tabular}{l|c|c|ccccccccccccccccccccc}
\specialrule{.15em}{.05em}{.05em}
\multirow{2}{*}{method} & \multirow{2}{*}{trainset} & \multirow{2}{*}{mAP (\%)} & \multicolumn{20}{c}{category} \\
&&& aero & bike & bird & boat & bottle & bus & car & cat & chair & cow & table & dog & horse & mbike & persn & plant & sheep & sofa & train & tv \\\specialrule{.15em}{.05em}{.05em}
Faster R-CNN \cite{SRenNIPS2015} & {\bf 07+12} & 76.4 & 79.8 & 80.7 & 76.2 & 68.3 & 55.9 & 85.1 & 85.3 & 89.8 & 56.7 & 87.8 & 69.4 & 88.3 & 88.9 & 80.9 & 78.4 & 41.7 & 78.6 & 79.8 & \underline{{\bf 85.3}} & 72.0 \\
ME R-CNN & {\bf 07+12} & \underline{{\bf 78.7}} & \underline{{\bf 81.2}} & \underline{{\bf 81.9}} & \underline{{\bf 78.0}} & \underline{{\bf 71.8}} & \underline{{\bf 65.0}} & \underline{{\bf 86.0}} & \underline{{\bf 87.5}} & \underline{{\bf 91.3}} & \underline{{\bf 61.0}} & \underline{{\bf 89.2}} & \underline{{\bf 69.9}} & \underline{{\bf 88.4}} & \underline{{\bf 90.1}} & \underline{{\bf 83.9}} & \underline{{\bf 81.4}} & \underline{{\bf 45.2}} & \underline{{\bf 81.0}} & \underline{{\bf 81.7}} & \underline{{\bf 85.3}} & \underline{{\bf 73.9}} \\\specialrule{.15em}{.05em}{.05em}
\end{tabular}
\end{center}
\vspace{-0.3cm}
\caption{{\bf VOC 2007 detection accuracy.} All methods use ResNet-101.}
\label{tab:voc07_resnet}
\end{table*}

\begin{table*}[t]
\setlength{\tabcolsep}{2.0pt}
\renewcommand{\arraystretch}{1.4}
\begin{center}
\begin{tabular}{l|c|c|ccccccccccccccccccccc}
\specialrule{.15em}{.05em}{.05em}
\multirow{2}{*}{method} & \multirow{2}{*}{trainset} & \multirow{2}{*}{mAP (\%)} & \multicolumn{20}{c}{category} \\
&&& aero & bike & bird & boat & bottle & bus & car & cat & chair & cow & table & dog & horse & mbike & persn & plant & sheep & sofa & train & tv \\\specialrule{.15em}{.05em}{.05em}
Fast R-CNN \cite{RGirshickICCV2015} & {\bf 12} & 65.7 & 80.3 & 74.7 & 66.9 & 46.9 & 37.7 & 73.9 & 68.6 & 87.7 & 41.7 & 71.1 & 51.1 & 86.0 & 77.8 & 79.8 & 69.8 & 32.1 & 65.5 & 63.8 & 76.4 & 61.7 \\
ME R-CNN$\star$ & {\bf 12} & {\bf 67.8} & 82.6 & 76.4 & 69.9 & 50.3 & 41.8 & 75.5 & 71.1 & 87.0 & 42.0 & 74.3 & 56.0 & 86.3 & 81.5 & 78.9 & 72.4 & 34.1 & 68.5 & 62.6 & 79.6 & 64.7 \\\hline
Faster R-CNN \cite{SRenNIPS2015} & {\bf 12} & 67.0 & 82.3 & 76.4 & 71.0 & 48.4 & 45.2 & 72.1 & 72.3 & 87.3 & 42.2 & 73.7 & 50.0 & 86.8 & 78.7 & 78.4 & 77.4 & 34.5 & 70.1 & 57.1 & 77.1 & 58.9 \\
ME R-CNN$\dagger$ & {\bf 12} & {\bf 69.2} & 81.2 & 75.7 & 71.2 & 51.1 & 47.8 & 73.3 & 74.6 & 88.1 & 46.9 & 76.4 & 52.9 & 87.1 & 81.7 & 81.4 & 78.8 & 38.4 & 72.9 & 60.0 & 78.4 & 66.9 \\\hline
Fast R-CNN \cite{RGirshickICCV2015} & {\bf 07++12} & 68.4 & 82.3 & 78.4 & 70.8 & 52.3 & 38.7 & 77.8 & 71.6 & \underline{{\bf 89.3}} & 44.2 & 73.0 & 55.0 & 87.5 & 80.5 & 80.8 & 72.0 & 35.1 & 68.3 & 65.7 & 80.4 & 64.2 \\
ME R-CNN$\ddagger$ & {\bf 07++12} & {\bf 70.7} & 84.0 & 79.8 & 72.4 & 54.9 & 43.3 & 78.4 & 74.7 & \underline{{\bf 89.3}} & 46.6 & 76.1 & \underline{{\bf 60.6}} & \underline{{\bf 87.8}} & \underline{{\bf 83.6}} & 82.1 & 74.8 & 39.4 & 70.6 & \underline{{\bf 65.7}} & \underline{{\bf 82.5}} & 67.9 \\\hline
Faster R-CNN \cite{SRenNIPS2015} & {\bf 07++12} & 70.4 & 84.9 & 79.8 & \underline{{\bf 74.3}} & 53.9 & 49.8 & 77.5 & 75.9 & 88.5 & 45.6 & 77.1 & 55.3 & 86.9 & 81.7 & 80.9 & 79.6 & 40.1 & 72.6 & 60.9 & 81.2 & 61.5 \\
ME R-CNN$\mathsection$ & {\bf 07++12} & \underline{{\bf 73.3}} & \underline{{\bf 85.4}} & \underline{{\bf 80.7}} & 74.0 & \underline{{\bf 58.3}} & \underline{{\bf 55.0}} & \underline{{\bf 79.7}} & \underline{{\bf 78.5}} & 88.6 & \underline{{\bf 52.9}} & \underline{{\bf 78.2}} & 57.8 & 87.7 & 83.3 & \underline{{\bf 83.7}} & \underline{{\bf 81.9}} & \underline{{\bf 50.6}} & \underline{{\bf 74.8}} & 62.4 & 81.8 & \underline{{\bf 69.8}} \\\specialrule{.15em}{.05em}{.05em}
\end{tabular}

\renewcommand{\arraystretch}{1.4}
\begin{tabular}{ll}
$\star$ \scriptsize{\texttt{http://host.robots.ox.ac.uk:8080/anonymous/69D0YS.html}} &
$\dagger$ \scriptsize{\texttt{http://host.robots.ox.ac.uk:8080/anonymous/O3RFBG.html}} \\
$\ddagger$ \scriptsize{\texttt{http://host.robots.ox.ac.uk:8080/anonymous/PLPKPU.html}} &
$\mathsection$ \scriptsize{\texttt{http://host.robots.ox.ac.uk:8080/anonymous/YTVCEH.html}}
\end{tabular}
\end{center}
\vspace{-0.3cm}
\caption{{\bf VOC 2012 detection accuracy.} All methods use VGG16.}
\label{tab:voc12}
\end{table*}

\begin{table*}[t]
\setlength{\tabcolsep}{2.0pt}
\renewcommand{\arraystretch}{1.4}
\begin{center}
\begin{tabular}{l|c|c|ccccccccccccccccccccc}
\specialrule{.15em}{.05em}{.05em}
\multirow{2}{*}{method} & \multirow{2}{*}{trainset} & \multirow{2}{*}{mAP (\%)} & \multicolumn{20}{c}{category} \\
&&& aero & bike & bird & boat & bottle & bus & car & cat & chair & cow & table & dog & horse & mbike & persn & plant & sheep & sofa & train & tv \\\specialrule{.15em}{.05em}{.05em}
Faster R-CNN \cite{SRenNIPS2015} & {\bf 07++12} & 73.8 & 86.5 & 81.6 & \underline{{\bf 77.2}} & 58.0 & 51.0 & 78.6 & 76.6 & \underline{{\bf 93.2}} & 48.6 & \underline{{\bf 80.4}} & 59.0 & \underline{{\bf 92.1}} & \underline{{\bf 85.3}} & 84.8 & 80.7 & 48.1 & 77.3 & 66.5 & \underline{{\bf 84.7}} & 65.6 \\
ME R-CNN$\star$ & {\bf 07++12} & \underline{{\bf 76.1}} & \underline{{\bf 87.1}} & \underline{{\bf 82.7}} & 76.3 & \underline{{\bf 62.5}} & \underline{{\bf 62.6}} & \underline{{\bf 81.7}} & \underline{{\bf 80.8}} & 90.6 & \underline{{\bf 54.8}} & 79.1 & \underline{{\bf 63.1}} & 89.6 & 84.4 & \underline{{\bf 85.4}} & \underline{{\bf 84.1}} & \underline{{\bf 55.0}} & \underline{{\bf 77.9}} & \underline{{\bf 67.1}} & 84.3 & \underline{{\bf 71.9}} \\\specialrule{.15em}{.05em}{.05em}
\end{tabular}

\renewcommand{\arraystretch}{1.4}
\begin{tabular}{ll}
$\star$ \scriptsize{\texttt{http://host.robots.ox.ac.uk:8080/anonymous/M9ZUJK.html}} & 
\end{tabular}
\end{center}
\vspace{-0.3cm}
\caption{{\bf VOC 2012 detection accuracy.} All methods use ResNet-101.}
\label{tab:voc12_resnet}
\end{table*}

PASCAL VOC datasets contain 20 object categories. VOC 07 dataset consists of 5k images in trainval set and 5k images in test set. VOC12 dataset has 10k images in trainval set and 10k images in test set. We use the standard metric for evaluating the detection accuracy for PASCAL VOC which is by taking a mean of average precision (mAP) over all object categories.

Table~\ref{tab:voc07} shows that, on VOC07, ME R-CNN provides improved detection accuracy in mAP than Fast/Faster RCNN when using VOC07 trainval set for training (69.0\% vs. 66.9\% and 70.6\% vs. 69.9\%, respectively). When using {\bf 07+12}, ME R-CNN outperforms both Fast R-CNN and Faster R-CNN by 2.2\% and 2.6\%, respectively (72.2\% vs. 70.0\% and 75.8\% vs. 73.2\%). VOC12 results are shown in Table~\ref{tab:voc12} where we observe consistent performance boost for ME R-CNN. In both cases of VOC12 trainval and 07++12, ME R-CNN outperforms both Fast/Faster R-CNN by at least 2.1\% mAP (2.9\% at most).

In table~\ref{tab:voc07_resnet} and~\ref{tab:voc12_resnet}, ME R-CNN shows a consistent performance boost when compared with ResNet-101 with Faster R-CNN on both VOC07 (78.7\% vs. 76.4\%) and VOC12 (76.1\% vs. 73.8\%). For this result, ME R-CNN was built on top of the ResNet-101 with Faster R-CNN architecture for fair comparison, also showing that the proposed architecture can effectively be combined with various types of object detection CNNs.

\subsection{MS COCO Results}
\label{ssec:COCOResults}

\begin{table*}[t]
\setlength{\tabcolsep}{15pt}
\renewcommand{\arraystretch}{1.4}
\begin{center}
\begin{tabular}{l|c|c|c|c|c|c }
\specialrule{.15em}{.05em}{.05em}
method & trainset & testset & mAP@.5 & Gain & mAP@[.5,.95] & Gain \\\specialrule{.15em}{.05em}{.05em}
Faster R-CNN \cite{SRenNIPS2015} & {\bf train} & {\bf val} & 41.5 & $\cdot$ & 21.2 & $\cdot$ \\
ME R-CNN & {\bf train} & {\bf val} & \underline{{\bf 43.6}} & +2.1 & \underline{{\bf 21.8}} & +0.6 \\\hline
Faster R-CNN \cite{SRenNIPS2015} & {\bf trainval} & {\bf test-dev} & 42.7 & $\cdot$ & 21.9 & $\cdot$ \\
ME R-CNN & {\bf trainval} & {\bf test-dev} & \underline{{\bf 45.3}} & +2.6 & \underline{{\bf 25.9}} & +4.0 \\\specialrule{.15em}{.05em}{.05em}
\end{tabular}
\end{center}
\vspace{-0.3cm}
\caption{{\bf MS COCO detection accuracy.} All methods use VGG16. Set key: {\bf train}: MS COCO train set, {\bf val}: MS COCO val set, {\bf trainval}: MS COCO train and val sets, {\bf test-dev}: MS COCO test-dev set.}
\label{tab:ms_coco}
\end{table*}

We evaluate ME R-CNN on MS COCO dataset and show the results in Table~\ref{tab:ms_coco}. The MS COCO dataset contains 80k, 40k and 20k samples in train, val, and test-dev sets, respectively. We compare ME R-CNN with the Faster R-CNN on two different evaluation settings: i) training the network using train set and testing on val set, and ii) training the network on train and val set and testing on test-dev set. We used two different standard metrics, which are mAP@.5 (PASCAL VOC metric) and mAP@[.5,.95] (MS COCO metric). The MS COCO metric (mAP@[.5,.95]) indicates the mAPs averaged for IOU$\in$[0.5:0.05:0.95]. Regardless of the which metric we choose to use, the ME R-CNN achieved consistent performance gain over the Faster R-CNN on both evaluation settings. When trained on train set, ME R-CNN outperforms the Faster R-CNN by 2.1 for mAP@.5 and 0.6 for mAP@[.5,.95]. When using train+val set for training, gains achieved by ME R-CNN over Faster R-CNN are 2.6 and 4.0 for the two metrics, respectively.


\subsection{Ablation Experiments}

The ablation experiments are conducted on PASCAL VOC07~\cite{MEveringhamIJCV2015}. In the ablation experiments, ME R-CNN model with Faster R-CNN and VGG16 is used. The model was trained on VOC07 trainval set and tested on VOC07 test set. \\

\noindent{\bf EAN vs. Hard-coded Assignment.}
We have verified the significance of employing the `learnable' EAN. Table \ref{tab:assigner} shows the detection accuracy of the ME R-CNN with and without EAN. As the architecture still requires an expert assigner to direct the RoIs to the experts even when EAN is not present, we have used a `hard-coded assignment' to take its place instead. For the hard-coded assignment case, the RoIs are assigned to the experts based on their aspect ratios which represent one of the three shape categories: horizontally elongated ({\bf H}), square-like ({\bf S}), and vertically elongated ({\bf V}). For training this model, the modified version of the 4-step alternating algorithm is used by updating the ME weights guided by the hard-coded assignments in step 2 and step 4. ME R-CNN with EAN module outperforms the hard-coded assignment case by 0.8 mAP which shows the effectiveness of having a learnable expert assigner in the multi-expert architecture. \\

\begin{table}[t]
\setlength{\tabcolsep}{21.0pt}
\renewcommand{\arraystretch}{1.4}
\begin{center}
\begin{tabular}{l|cc}
\specialrule{.15em}{.05em}{.05em}
Assigner & EAN & Hard-coded assignment \\\specialrule{.15em}{.05em}{.05em}
mAP (\%) & {\bf 70.6} & 69.8 \\\specialrule{.15em}{.05em}{.05em}
\end{tabular}
\end{center}
\vspace{-0.3cm}
\caption{{\bf Effectiveness of learning RoI-expert relationship.} RoI-expert relationship is ``learned'' in EAN whereas RoIs are designated to the experts based on a pre-defined criteria in the hard-coded assignment.}
\label{tab:assigner}
\end{table}

\noindent{\bf Functionality of Experts.} Table \ref{tab:me_functionality} shows how the functionality of each expert has changed after the overall training has been carried out. As mentioned in `Initialize ME \& Conv-L' in Subsection \ref{ssec:co-training}, ME weights are initialized by directing the RoIs to the experts by their aspect ratio-based shape categories ({\bf H}, {\bf S}, or {\bf V}). The table shows that the newly trained experts, with the help of `learnable' EAN, is less biased towards the aspect ratio of the RoIs. 

The detection performance of each expert on the whole dataset (VOC07 test set) only reaches to approximately 66\% mAP which is lower than the performance of the Faster R-CNN. However, one can notice that when these multiple experts are exploited together, the performance can be boosted up to 70.6\% (0.7\% better than Faster R-CNN).

\begin{figure*}[t!]
\begin{minipage}[b]{1.0\linewidth}
  \centering
  \centerline{\includegraphics[width=\textwidth,trim=0mm 0mm 0mm 0mm]{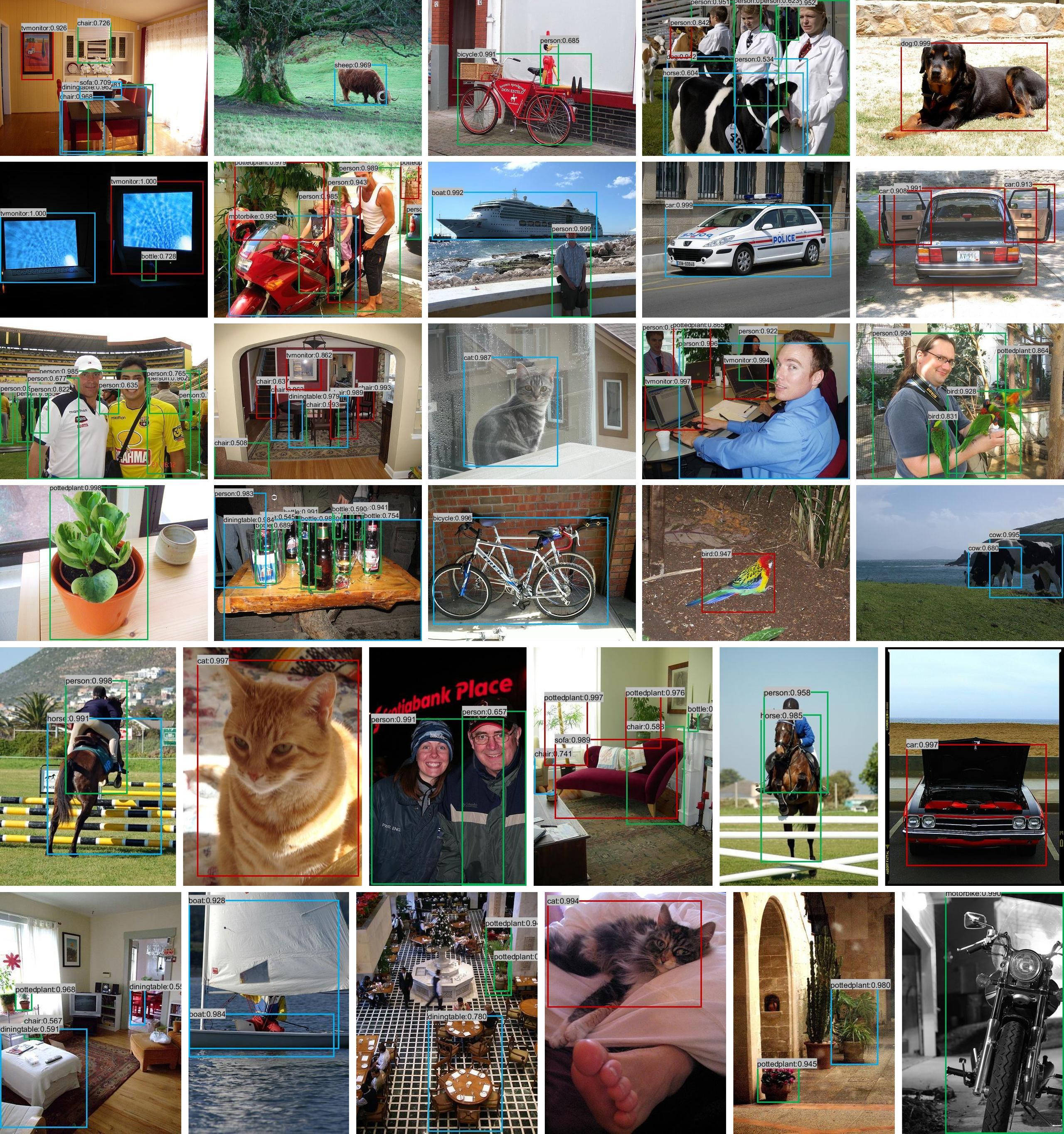}}
\end{minipage}
\caption{{\bf Example object detection results using ME R-CNN.} The detection results from the experts 1, 2, and 3 are depicted in red, blue, and green bounding boxes, respectively. }
\label{fig:examples}
\end{figure*}

We can also observe that the functionality change in the ME, made possible by the co-training of ME and EAN, eventually provided extra room for noticeable performance increase over the hard-coded assignment case (Table \ref{tab:assigner}). Figure \ref{fig:examples} depicts example object detection results acquired by different experts in our final version of ME R-CNN. \\

\begin{table*}[t]
\setlength{\tabcolsep}{14pt}
\renewcommand{\arraystretch}{1.4}
\begin{center}
\begin{tabular}{c|ccc||c}
\specialrule{.15em}{.05em}{.05em}
\multirow{2}{*}{Expert \#} & \multicolumn{3}{c||}{Detection Bounding Boxes} & \multirow{2}{*}{mAP (\%)} \\
& $w > \sqrt{2}h$ & $\sqrt{2}h \geq w \geq \frac{1}{\sqrt{2}}h$ & $w < \frac{1}{\sqrt{2}}h$ & \\\specialrule{.15em}{.05em}{.05em}
1 & 46.8\% (83.8\%) & 24.4\% (16.2\%) & 28.8\% (0.0\%) & 66.1 \\
2 & 33.4\% (24.7\%) & 44.2\% (50.2\%) & 22.4\% (25.1\%) & 66.7 \\
3 & 7.5\% (0.0\%) & 35.6\% (17.7\%) & 56.9\% (82.3\%) & 66.4 \\\specialrule{.15em}{.05em}{.05em}
\end{tabular}
\end{center}
\vspace{-0.3cm}
\caption{{\bf The change in the functionality of the experts with EAN.} Table shows the RoI-expert distribution after the overall training along with the final detection accuracy (mAP) for each expert. Note that the final distributions have drastically changed from the aspect ratio-based initialization (in parentheses). $w$ and $h$ indicate the width and the height of RoI, respectively.}
\label{tab:me_functionality}
\end{table*}

\noindent{\bf Co-training of ME, EAN, \& Conv-L.}
We have conducted an experiment to validate the effectiveness of co-training (simultaneous training) the three major components (ME, EAN, and Conv-L) of ME R-CNN. These three components affect each other which makes the training challenging as mentioned in \ref{ssec:co-training}. We compare our training strategy tailored for ME R-CNN (Algorithm 1) with the two baselines which do not perform any co-training for the three components. The first baseline (denoted as `Remove 2c' in Table \ref{tab:co-learning}) is implemented by simply removing step 2c from Algorithm 1. In this scenario, ME and EAN are trained twice during the entire training process. The second baseline (denoted as `Replace 2c') is implemented by replacing step 2c by alternating update of ME and EAN. ME and EAN are trained three times for this baseline. For both of the baselines, the last step which Conv-L is being optimized is step 2a in Algorithm 1. This is because, without using the strategy of co-training of the three components, Conv-L needs to be fixed. As can be seen in Table \ref{tab:co-learning}, having the three components trained simultaneously outperforms the baseline cases.\medskip

\begin{table}[t]
\setlength{\tabcolsep}{18.0pt}
\renewcommand{\arraystretch}{1.4}
\begin{center}
\begin{tabular}{c|c|c}
\specialrule{.15em}{.05em}{.05em}
Optimization Strategy & Co-training? & mAP (\%) \\\specialrule{.15em}{.05em}{.05em}
Remove 2c & No & 68.8 \\
Replace 2c & No & 69.3 \\
Algorithm 1 & Yes & {\bf 70.6} \\\specialrule{.15em}{.05em}{.05em}
\end{tabular}
\end{center}
\vspace{-0.3cm}
\caption{{\bf Effectiveness of co-training of EAN, ME, \& conv-L.} `2c' refers to Step 2c in Algorithm 1. This is the only step where co-training of the three components takes place.}
\label{tab:co-learning}
\end{table}

\noindent{\bf EAN Labels Presetting.} To understand how EAN label presetting affects object detection accuracy, we compared three RoI-expert assignment presetting cases based on various RoI characteristics: aspect ratio, RoI size, and object category. In each case, three experts were employed to make a fair comparison.

When size is used as the presetting criteria, we manually defined three different categories: {\it small}, {\it medium}, and {\it large}. All the RoIs bigger than 110$^2$ are assigned as {\it large}, and the other RoIs are assigned as {\it small}. All RoIs that fall between 55$^2$ and 205$^2$ are assigned as {\it medium}. Similar to aspect-ratio-based presetting criteria, one RoI can be assigned to either one or two categories.

When the presetting criteria is defined based on object category, the RoIs are grouped to be semantically similar as follows:
\begin{itemize}
    \item {\it Vehicl}e: Aeroplane, Bicycle, Boat, Bus, Car, Motorbike, Train
    \item {\it Animal}: Bird, Cat, Cow, Dog, Horse, Person, Sheep
    \item {\it Other}: Bottle, Chair, Diningtable, Pottedplant, Sofa, TVmonitor
\end{itemize}
RoIs having less than 0.1 IOU overlap with any object bounding box are not used in EAN training.

\begin{figure}[t]
\begin{minipage}[b]{1.0\linewidth}
  \centering
  \centerline{\includegraphics[width=\textwidth,trim=5mm 5mm 5mm 5mm,clip]{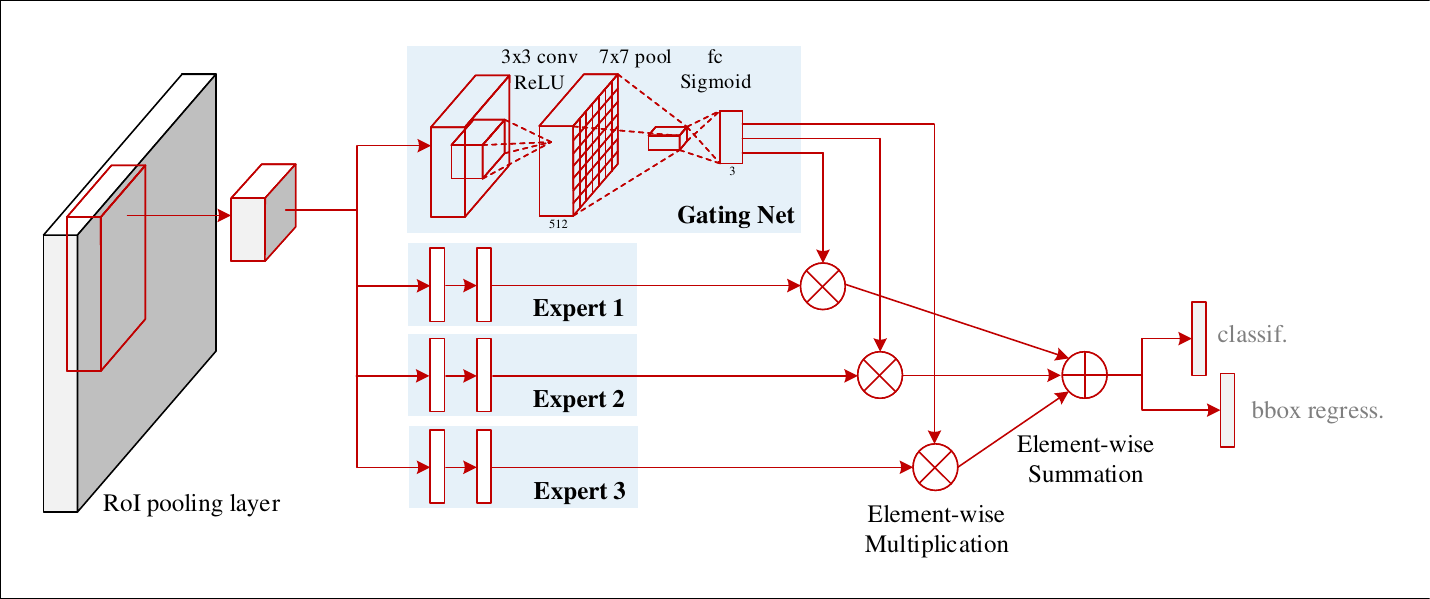}}
\end{minipage}
\caption{{\bf R-CNN architecture adopting mixture-of-expert in a conventional manner.} To provide fair grounds for comparison between this model and ME R-CNN with respect to the network size, the number of experts has been set to three, and the gating network is implemented to match the EAN architecture.}
\label{fig:moe_rcnn}
\end{figure}

Table~\ref{tab:ean_init} compares the object detection accuracy of the three presetting strategies. Presetting based on RoI's aspect ratio shows the highest accuracy, but the differences compared to the other two are marginal ($\sim$1\%). We observe that the accuracy is not highly dependent upon the EAN label presetting strategy.\medskip

\begin{table}[t]
\setlength{\tabcolsep}{11.9pt}
\renewcommand{\arraystretch}{1.4}
\begin{center}
\begin{tabular}{c|ccc}
\specialrule{.15em}{.05em}{.05em}
Presetting & Aspect Ratio & RoI Size & Object Category \\\specialrule{.15em}{.05em}{.05em}
mAP (\%) & {\bf 70.6} & 69.6 & 69.9 \\\specialrule{.15em}{.05em}{.05em}
\end{tabular}
\end{center}
\vspace{-0.3cm}
\caption{{\bf Comparison of three EAN label presetting.}}
\label{tab:ean_init}
\end{table}

\begin{table}[t]
\setlength{\tabcolsep}{9.3pt}
\renewcommand{\arraystretch}{1.4}
\begin{center}
\begin{tabular}{c|c|c|c}
\specialrule{.15em}{.05em}{.05em}
\multicolumn{2}{c|}{Conventional} & \multirow{2}{*}{mAP (\%)} & \multirow{2}{*}{Test time (sec)} \\\cline{1-2}
ImageNet Pretrain & Gating Net & & \\\specialrule{.15em}{.05em}{.05em}
& & 68.7 & 0.099 \\
& $\surd$ & 68.9 & 0.106 \\
$\surd$ & & 69.6 & 0.099 \\
$\surd$ & $\surd$ & 69.7 & 0.106 \\\hline\hline
\multicolumn{2}{c|}{ME R-CNN} & {\bf 70.6} & {\bf 0.075} \\\specialrule{.15em}{.05em}{.05em}
\end{tabular}
\end{center}
\vspace{-0.3cm}
\caption{{\bf Performance comparison of various models adopting mixture-of-expert in a conventional way with respect to object detection accuracy (mAP) and test time.} We compared four conventional approaches that differ depending on the use of finetuning from ImageNet pretrained network and the use of the gating network.}
\label{tab:mixture-of-expert}
\end{table}

\noindent{\bf Comparison with CNNs Adopting Mixture-of-Expert in a Conventional Way.} We compare ME R-CNN with CNNs that adopt the mixture-of-expert in a conventional manner. This comparison is made to verify the effectiveness of the proposed approach which assigns each RoI into its most appropriate expert instead of utilizing all the experts at once. Figure~\ref{fig:mixture-of-expert} shows the architectural differences between our approach and the conventional approaches when adopting the mixture-of-expert concept into a CNN architecture.

As a representative conventional architecture, we implemented R-CNN with three experts and a gating network tied in a conventional manner as shown in Figure~\ref{fig:moe_rcnn}. For the gating network, we use the same architecture (single convolutional layer, single pooling layer, and single fully-connected layer) as the EAN in ME R-CNN so that this conventional model has the same number of weights as ME R-CNN. This network is also optimized using 4-step alternating algorithm in exactly the same order. In the second step, each expert is optimized by minimizing its own softmax loss for classification and L1smooth loss for bounding box regression, i.e., six losses are imposed. Accordingly, we prepare three different training batches so that each expert is trained with a different batch. Meanwhile, in the fourth step only one set of losses (classification and bounding box regression) is used to optimize all three experts. The gating network was trained only in the fourth step.

A common goal for having multiple experts is to be able to equip each expert module with its own unique expertise and thus leverage such capability for overall performance increase. However, finetuning from the same ImageNet-pretrained network (i.e., same initialization) is likely to lead all experts to have similar expertise which conflicts with the original goal. Therefore, we have trained this conventional model in two different ways, with and without using the pretraining network. To observe the effectiveness of the gating network, we compared Figure~\ref{fig:moe_rcnn} with the model without the gating network. In Table~\ref{tab:mixture-of-expert}, four combinations of the conventional mixture-of-expert approaches are compared with ME R-CNN with respect to mAP and test time. We can observe that the object detection accuracy underperforms when the finetuning strategy was omitted, which shows that our original anticipation to differentiate the experts' expertise was not achieved. It is also shown that the gating network improves performance marginally. Above all, ME R-CNN provides better mAP and speed than any of the conventional mixture-of-expert approaches.

\subsection{Timing}
To analyze the computational overhead of exploiting ``multiple experts'', we compare the train/test time of ME R-CNN with the Faster R-CNN. This analysis was conducted using NVidia Titan XP GPU.\medskip

\noindent{\bf Inference.} 
While Faster R-CNN takes 0.07 sec/image, ME R-CNN takes 0.075 sec/image. Using multiple experts brings almost no overhead because the number of RoIs do not change compared to the Faster R-CNN case, and only one of the experts is being activated for each RoI.\medskip

\noindent{\bf Training.} Training ME R-CNN (16.9 hrs) requires almost twice as much time when compared to the case of Faster R-CNN (8.15 hrs). Several recently introduced measures such as multi-GPU parallel computing or enlarging mini-batch size~\cite{KHeCVPR2015,PGoyalArxiv2017} can be taken into consideration to reduce the overall training time.

\section{Future Works}
\label{sec:future}
In this paper, we have focused on showing that ME R-CNN architecture can boost the object detection performance when integrated with the baseline R-CNNs. Fast R-CNN and Faster R-CNN, which are two of the most widely used object detection networks, were selected to demonstrate the effectiveness.

In the future, we will focus on producing the state-of-the art performance in benchmark datasets (PASCAL VOC and MS COCO) by incorporating additional processings (refered to as `adding bells and whistles' in \cite{AShrivastavaCVPR2016}) such as multi-scale training/testing (MS)~\cite{RGirshickICCV2015,KHeECCV2014}, online hard example mining (OHEM)~\cite{AShrivastavaCVPR2016}, iterative bounding
box regression~\cite{SGidarisICCV2015}, global context (CXT)~\cite{SGidarisICCV2015,KHeCVPR2016,JLiTMultimedia2017}, ensemble of classifiers (ENS)~\cite{KHeCVPR2016,HLeeWACV2016}, integrating with image classification output~\cite{YCaoICIP2017}, and feature pyramid network (FPN)~\cite{TLinCVPR2017}. We also plan to incorporate multi-expert into other CNN-based detection architecture such as SSDs~\cite{WLiuECCV2016,CFuArxiv2017}, YOLOs~\cite{JRedmonCVPR2016,JRedmonCVPR2017,JRedmonArxiv2018}, R-FCN~\cite{JDaiNIPS2016}, and RetinaNet~\cite{TLinICCV2017}, which has recently been introduced.\medskip

\begin{table*}[t]
\setlength{\tabcolsep}{3.1pt}
\renewcommand{\arraystretch}{1.4}
\begin{center}
\scriptsize{
\begin{tabular}{c|c|c|c|c|c|c}
\specialrule{.15em}{.05em}{.05em}
\multirow{2}{*}{Rank} & \multirow{2}{*}{Method} & \multicolumn{2}{c|}{Backbone} & \multicolumn{3}{c}{Adding Bells and Whistles} \\\cline{3-7}
& & Img. Classif. & Obj. Det. & Plug-In & Boosting Strategy & More Info.\\
\specialrule{.15em}{.05em}{.05em}
1 & MegDet~\cite{CPengCVPR2018} & ResNeXt-152~\cite{SXieCVPR2017} & Faster & FPN, GCN~\cite{CPengCVPR2017}, RoI Align~\cite{KHeICCV2017} & {\bf Large Batch}, OHEM, CXT, MS, ENS & Segmentation~\cite{JMaoCVPR2017} \\
2 & PANet~\cite{SLiuCVPR2018} & ResNet-101 & Faster & {\bf Adaptive FeatPool}, FPN, RoI Align~\cite{KHeICCV2017} & MS, ENS & Pixel Label~\cite{KHeICCV2017} \\
3 & MSRA & Xception~\cite{FCholletCVPR2017} & Faster & {\bf Novel RoI Pool~\cite{JDaiCVPR2017}}, FPN, SoftNMS~\cite{NBodlaICCV2017} & ENS & $\cdot$ \\
4 & Mask R-CNN~\cite{KHeICCV2017} & ResNeXt-152~\cite{SXieCVPR2017} & Faster & {\bf RoI Align}, RPN & MS, ENS & Pixel Label~\cite{KHeICCV2017}, ImageNet-5K \\
\vdots & \vdots & \vdots & \vdots & \vdots & \vdots & \vdots \\ 
\specialrule{.15em}{.05em}{.05em}
\end{tabular}
}
\end{center}
\vspace{-0.3cm}
\caption{{\bf Leaderboard of MS COCO object detection competition (As of June 14, 2019).} The major contribution of each method is shown in bold. For the image classification backbone, ResNeXt and Xception were devised based on the ResNet architecture. All the acronyms in the table have previously been defined.}
\label{tab:top_ranked_methods}
\end{table*}

\noindent{\bf State-of-the-art Object Detection Methods.} On the leaderboard of the MS COCO object detection competition~\cite{MSCOCOleaderboard}, the top ranked methods provide much higher object detection accuracy compared to the accuracy achieved with our approach.\footnote{Details on the architectures used for the top ranked methods in the PASCAL VOC Competition are unavailable.} The top four methods~\cite{CPengCVPR2018,SLiuCVPR2018,JDaiCVPR2017,KHeICCV2017} achieve the state-of-the-art accuracy by adding multiple plug-in modules, using performance boosting strategies, and/or providing more information to the backbone combining Faster R-CNN with one of the ResNet variations (ResNet, ResNeXt or Xception). Table~\ref{tab:top_ranked_methods} shows the bells and whistles used to achieve the state-of-the-art accuracy of each method. Following this trend, we also experimented by adding our ME module to the backbone combining Faster R-CNN and ResNet.

\section{Conclusion}
\label{sec:Conclusion}
We introduced ME R-CNN which uses multiple experts (ME) in place of a conventional single classifier incorporated in CNN-based object detection architecture. Having ME is found to be advantageous as each expert is learned to specialize in a certain type of RoIs, considering the fact that RoIs are manifested in various appearance caused by different shapes, poses, and viewing angles. To optimize the ME usage, we have introduced expert assignment network (EAN) which automatically learns the RoI-expert relationship. We have introduced a practical training strategy to better handle the challenging task of optimizing the complex architecture which contains ME, EAN, and a shared convolutional network. With benefits of the novel components, ME R-CNN proves its effectiveness in consistently enhancing the detection accuracy in PASCAL VOC 07, 12, and MS COCO datasets over the baseline methods.


%

\ifCLASSOPTIONcaptionsoff
  \newpage
\fi



\bibliographystyle{IEEEtran}
\bibliography{references.bib}
\end{document}